\documentclass[10pt,conference]{IEEEtran}
\IEEEoverridecommandlockouts

\usepackage{cite}
\usepackage{amsmath,amssymb,amsfonts}
\usepackage{algorithmic}
\usepackage{graphicx}
\usepackage{textcomp}
\usepackage{xcolor}
\usepackage{amsthm}
\usepackage{subfigure}
\usepackage{colortbl}
\usepackage{soul}  
\usepackage{enumitem}
\usepackage{wrapfig}
\usepackage{listings}
\usepackage{xcolor}
\definecolor{skyblue}{rgb}{0., 0.72, 0.92}
\usepackage{amsmath, amsfonts}
\usepackage[outline]{contour}
\usepackage[linesnumbered,ruled,vlined]{algorithm2e}
\SetKwInOut{Input}{Input}
\SetKwInOut{Output}{Output}

\SetCommentSty{mycommfont}

\usepackage{multirow}
\usepackage{booktabs}

\theoremstyle{definition}

\definecolor{lightgray}{rgb}{0.78, 0.78, 0.78}
\sethlcolor{lightgray}  
\definecolor{answercolor}{RGB}{240, 240, 240}

\def\BibTeX{{\rm B\kern-.05em{\sc i\kern-.025em b}\kern-.08em
    T\kern-.1667em\lower.7ex\hbox{E}\kern-.125emX}}

\begin{document}

\title{Evaluating the Robustness of Test Selection Methods for Deep Neural Networks}

\author{\IEEEauthorblockN{Qiang Hu\IEEEauthorrefmark{1},
Yuejun Guo\IEEEauthorrefmark{2}, 
Xiaofei Xie\IEEEauthorrefmark{3}, 
Maxime Cordy\IEEEauthorrefmark{1},
Wei Ma\IEEEauthorrefmark{4},
Mike Papadakis\IEEEauthorrefmark{1} and
Yves Le Traon\IEEEauthorrefmark{1}}
\IEEEauthorblockA{\IEEEauthorrefmark{1}University of Luxembourg, Luxembourg\\
\IEEEauthorrefmark{2}Luxembourg Institute of Science and Technology, Luxembourg \\
\IEEEauthorrefmark{3}Singapore Management University, Singapore  \\
\IEEEauthorrefmark{4}Nanyang Technological University, Singapore
}}

\maketitle

\begin{abstract}
Testing deep learning-based systems is crucial but challenging due to the required time and labor for labeling collected raw data. To alleviate the labeling effort, multiple test selection methods have been proposed where only a subset of test data needs to be labeled while satisfying testing requirements. However, we observe that such methods with reported promising results are only evaluated under simple scenarios, e.g., testing on original test data. This brings a question to us: are they always reliable? In this paper, we explore when and to what extent test selection methods fail for testing. Specifically, first, we identify potential pitfalls of 11 selection methods from top-tier venues based on their construction. Second, we conduct a study on five datasets with two model architectures per dataset to empirically confirm the existence of these pitfalls. Furthermore, we demonstrate how pitfalls can break the reliability of these methods. Concretely, methods for fault detection suffer from test data that are: 1) correctly classified but uncertain, or 2) misclassified but confident. Remarkably, the test relative coverage achieved by such methods drops by up to 86.85\%. On the other hand, methods for performance estimation are sensitive to the choice of intermediate-layer output. The effectiveness of such methods can be even worse than random selection when using an inappropriate layer.

\end{abstract}

\maketitle

\section{Introduction}

Deep learning (DL) has been a critical technique in our daily life and multiple DL-based systems have been deployed, for example, face recognition systems~\cite{hu2015face}, chatbots~\cite{wu2019deep}, and self-driving cars~\cite{badue2021self}. Similar to conventional software systems that need to be systematically tested, testing DL-based systems is crucial to ensure that they meet the performance expectation and user requirements. Generally, the key component of DL-based systems is the deep neural network (DNN). Thus, the straightforward way of testing such systems is to evaluate their embedded pre-trained DNNs. 

In a nutshell, DNN follows a data-driven paradigm and learns the inference logic from the training data, then makes decisions for new data. High-quality training data are important for preparing DNNs, and diverse test data with their corresponding labels are necessary for testing the performance of pre-trained DNNs. However, labeled test data are difficult to obtain due to the heavy labeling process, intensive human effort, and required domain knowledge. How to efficiently and fully test DNNs with fewer labeled test data is a challenging problem and becomes a hot research direction recently. 

One promising solution to tackle this labeling problem is test selection for deep learning~\cite{ma2021test} -- select and only label a subset of data from the entire test set. Multiple test selection methods have been proposed mainly by the software engineering community. Those methods can be roughly divided into two categories, 1) test selection for quick fault detection, and 2) test selection for model performance estimation. For simplicity, we call these two methods \textbf{fault detection method} and \textbf{performance estimation method}. Fault detection methods tend to identify the test data that have a higher probability to be misclassified by the model as faults. This kind of data can be further used for repairing the model (mainly by retraining) to enhance its performance. Performance estimation methods try to select a small size of data that can represent the entire test set. The model performance on the entire test set can be then approximated by using this subset. Here, robust means the capability of test selection methods to preserve their effectiveness on unusual (often, altered) data. 

Similar to DL targeted test generation techniques whose reliability needs to be carefully studied~\cite{riccio2022and} before real usage. The reliability of test selection methods also requires systematic exploration to check when they could fail, and more importantly, how harmful the failure could bring to the users. Otherwise, under the risky test selection methods, the testing requirements might not be satisfied and the labeling budget will be wasted. For example, fault detection methods rely on the assumption -- there is a strong correlation between the probability of misclassification and the uncertainty of the data. They believe the data with high uncertainty (i.e., the model has low confidence in the data) should be misclassified by the model. Here, uncertainty is the proneness of input data to be mishandled by a model, which can be measured in different ways, e.g., entropy of the output probabilities. Confidence refers to how much the model believes it has made the correct decision. In classification models, this is the class probability as output by the softmax layer. However, this assumption leads the selection process to ignore faults with low uncertainty. Thus, when facing a large number of low uncertain data, the effectiveness of fault detection methods may degrade a lot. We conjecture that \textit{existing test selection methods are not robust.} It is necessary to know what are their limitations for reliable usage.

In this paper, we study the above problems and focus on exploring when and to what extent the test selection methods fail. Concretely, first, we collect and make a short survey of existing fault detection methods and performance estimation methods. In total, there are eight methods for fault detection and three methods for performance estimation (including random selection). Second, we analyze the potential pitfalls that could make these methods fail based on their design strategies. After that, we conduct an empirical study to reveal whether these issues exist and if yes, how harmful they are. Here, we design test generation methods to trigger these issues. Finally, based on our findings, we provide guidelines on how to robustly evaluate test selection methods when developing new ones. In total, our study covers five widely studied datasets (e.g., Traffic-sign, CIFAR100) with two popular model architectures (e.g., ResNet, DenseNet) for each dataset. Based on our study, we found:

\begin{itemize}[leftmargin=*]

\item All fault detection methods are fooled by two types of test data, 1) correctly classified but with high uncertainty data, and 2) wrongly classified but with low uncertainty data. On average, the effectiveness of test selection methods drops 52.49\% and 39.80\% when facing these two types of test data, respectively.

\item These two types of test data have a huge negative impact on the selection-guided model repair (i.e. retraining). When facing them, even random selection can achieve better repair results than well-designed methods like PRIMA, TestRank, and DSA. Specifically, 55 out
of 80 repaired models have accuracy degradation
when dealing with correctly classified but uncertain data.

\item The effectiveness of existing performance estimation methods highly depends on the choice of output from the intermediate layer. They can only outperform the random selection with handpicked layers. 
\end{itemize}

To summarize, the main contributions of this paper are:
\begin{enumerate}[leftmargin=*]
\item This is the first work that explores the limitations of test selection methods for DNNs.
\item We reveal that existing 1) fault detection methods suffer from the correctly (wrongly) classified but uncertain (with high confidence) test data, and 2) performance estimation methods are sensitive to the choice of intermediate layers.

\end{enumerate}

\section{Background and Related Work}
\label{sec:background}

\subsection{Deep Learning Testing}

Roughly speaking, the objectives of testing DNNs mainly lie in two parts, 1) testing the natural robustness of DNNs~\cite{hendrycks2019benchmarking,ovadia2019can,chen2021mandoline} for ensuring their performance facing data with diverse distributions (including the same distribution as training data), and 2) testing the adversarial robustness of DNNs~\cite{carlini2017towards,carlini2019evaluating} for potential security issues (e.g., adversarial attack). For the natural robustness testing, developers need to prepare as many as possible test data that follow different data distributions, and then assess DNN models accordingly. For adversarial robustness testing, generating adversarial examples and using them to test the models is a common way. Utilizing adversarial robustness certification techniques~\cite{gehr2018ai2,singh2018fast} to strictly verify the model robustness is another way. For both types of DNN testing, labeled test data are necessarily needed. In this paper, we focus on the first objective of DNN testing and target the test data selection for efficient deep learning testing. 

\subsection{Test Selection for DNNs}

Test selection methods are proposed for efficiently testing DNNs without heavy labeling effort. There are two types of test selection methods, the first one tries to select data that are most likely misclassified by the model~\cite{kim2019guiding,feng2020deepgini,9286133,ma2021test,NEURIPS2021_ae785101,9402064,9793868}, and the second selects data~\cite{li2019boosting, chen2020practical} to approximate the model performance on the entire test set. We introduce methods from each type in Section~\ref{sec:mertics}. In this paper, instead of proposing new test selection methods, we reveal and study when existing test selection methods could fail. 

\subsection{Empirical study on Deep Learning Testing}

Similar to this paper, many works~\cite{yang2022revisiting} conducted empirical studies to investigate the usefulness of existing deep learning testing techniques. Yan \emph{et al.}~\cite{yan2020correlations} and Dong \emph{et al.}~\cite{dong2020empirical} studied the correlations between the neural coverage~\cite{deepxplore2017,ma2018deepgauge} criteria (a type of deep learning testing criteria) and the model quality (mainly focus on the adversarial robustness of models). Jahangirova \emph{et al.}~\cite{riccio2021deepmetis} empirically evaluated the effectiveness of mutation testing for DNNs by using different mutation operators. Hu \emph{et al.}~\cite{hu2022empirical} specifically studied how the test selection methods guided model retraining performance when facing data with different types of distribution shifts. Two works~\cite{ma2021test,weiss2022simple} studied the effectiveness of fault detection target test selection methods, and they found that simple methods perform well in terms of fault detection and model retraining. Different from existing studies, our work focuses on two types of test selection methods and does not only consider simple test data. We build more complex testing scenarios to challenge the test selection methods and reveal their weak point.

\section{Test Selection Methods Analysis}
\label{sec:mertics}

\begin{table*}[t]
\centering
\caption{Summary of Test Selection Methods.} 
\label{tab:ts_methods}
\resizebox{\textwidth}{!}
{
\begin{tabular}{lllll}
\hline
\textbf{Type} & \textbf{Method Name} & \textbf{Feature Used} & \textbf{Target Task} & \textbf{Venue} \\ \hline
\multirow{8}{*}{\textbf{Fault detection}} & Distance-based surprise adequacy (DSA)~\cite{kim2019guiding} & Intermediate output & \begin{tabular}[c]{@{}l@{}}Classification\\ Regression\end{tabular} & ICSE 2019  \\
 
 & DeepGini~\cite{feng2020deepgini} & Final output & Classification & ISSTA 2020  \\
 & Multiple-boundary clustering and prioritization (MCP)~\cite{9286133} & Final output & Classification & ASE 2020  \\
 & Monte Carlo dropout uncertainty (MC)~\cite{ma2021test} & Final output & Classification & TOSEM  \\
 & Maximum probability (MaxP)~\cite{ma2021test} & Final output & Classification & TOSEM  \\
 & TestRank~\cite{NEURIPS2021_ae785101} & \begin{tabular}[c]{@{}l@{}}Logits output\\ + graph model\end{tabular} & Classification & NeurIPS 2021  \\
 & PRIMA~\cite{9402064} & \begin{tabular}[c]{@{}l@{}}Final output\\ + ranking model\end{tabular} & \begin{tabular}[c]{@{}l@{}}Classification\\ Regression\end{tabular} & ICSE 2021  \\
 & Adaptive test selection (ATS)~\cite{9793868} & Final output & Classification & ICSE 2022 \\ \hline
\multirow{2}{*}{\textbf{Performance estimation}} & Cross entropy-based sampling (CES)~\cite{li2019boosting} & Intermediate output & Classification & ESEC/FSE 2019  \\
 & Practical accuracy estimation (PACE)~\cite{chen2020practical} & \begin{tabular}[c]{@{}l@{}}Intermediate output\\ + clustering model\end{tabular} & \begin{tabular}[c]{@{}l@{}}Classification\\ Regression\end{tabular} & TOSEM  \\ \hline
\end{tabular}
}
\end{table*}

In total, We collect 10 representative test selection methods plus the random selection based on existing works~\cite{ma2021test, weiss2022simple} and our review. Table~\ref{tab:ts_methods} presents a brief description of methods, data features used for conducting selection, target tasks, and the venue where a method comes from. According to the purpose of use, existing test selection methods can be classified into two types. The first one is test selection for fault detection (fault detection method), and the second is test selection for model performance estimation (performance estimation method). In this work, we focus on the classification task since most of the methods (7 out of 10) are specifically designed for this task.

Let $f:\mathcal{X}\rightarrow\mathcal{Y}$ be a trained DNN model that maps input data $\mathcal{X}$ into the target space $\mathcal{Y}$. $x\in\mathcal{X}$ and $y$ denote a given test sample and its true label, respectively. The model $f$ assigns the label $\widehat{y}$ to $x$ based on its learned knowledge from the training set. Let $p_{y_i}\left(x\right)$ be the likelihood of $x$ belonging to class $y_i\in\mathcal{Y}$, $\widehat{y}=y_i$ if $p_{y_i}$ is the maximum over all possible classes in $\mathcal{Y}$.

\subsection{Fault Detection Methods}

In this paper, \textit{fault} refers to perceptible discordance between the model decision and the expected outcome. In classification tasks that we study in this work, a fault is revealed by an incorrect label. Given a set of unlabeled test data and a DNN model, fault detection methods identify test data (faults) that are likely to be wrongly predicted by the model. Generally, compared to the correctly predicted data, people are more interested in fault data that are useful to analyze the weak/blind point of the DNN model and further enhance the model. In practice, the found faults are usually utilized as a \textit{patch} to repair (via model retraining) DNN models. 

\textbf{Distance-based surprise adequacy (DSA)}~\cite{kim2019guiding} is the very first method proposed to select test data. The main idea of DSA is to measure the difference in activation traces between a given test sample and the training set.

\begin{equation}
\label{equ:dsa}
DSA\left(x\right)=\frac{dist_a}{dist_b}
\end{equation}
where $dist_a$ is the Euclidean distance between $x$ and its closest neighbor from the same class $\widehat{y}$ and $dist_b$ refers to the neighbor from a different class. A sample with a high difference can better reveal faults.

\textbf{DeepGini}~\cite{feng2020deepgini} quantifies the uncertainty of $x$ for $f$:
\begin{equation}
\label{equ:gini}
Gini\left(x\right)=1-\sum_{y_i\in\mathcal{Y}}p_{y_i}\left(x\right)
\end{equation}
Test samples with high uncertainties are considered as insufficiently learned by $f$ and selected to detect faults.

\textbf{Multiple-boundary clustering and prioritization (MCP)}~\cite{9286133} first performs the so-called boundary area clustering that divides data into different boundary areas (confusion areas of every two classes). Based on the prediction, MCP assigns a priority:
\begin{equation}
    MCP\left(x\right)=\frac{p_{\widehat{y}}}{p_{y_s}}
\end{equation}
where $y_s$ is the predicted second most likely class. Next, from each boundary area, MCP evenly selects data with the minimum priority.

\textbf{Monte Carlo dropout uncertainty (MC)}~\cite{ma2021test} takes advantage of the Monte Carlo dropout technique $f$ to obtain multiple predictions and then calculate the uncertainty score of the data. There are several variants of MC, we consider the famous one~\cite{hu2021towards,siddhant2018deep} in which the uncertainty is calculated by:
\begin{equation}
\label{equ:dropout}
MC\left(x\right)=1-\frac{\mid\left\{i\mid\widehat{y_i}=
mode\left\{\widehat{y_j},1\leq j\leq N\right\}\right\}\mid}{N}
\end{equation}
where $N$ is the number of applying the dropout to obtain predictions. $mode\left(.\right)$ is the function of identifying the predicted class that appears most often.

\textbf{Maximum probability (MaxP)}~\cite{ma2021test} considers $p_{\widehat{y}}\left(x\right)$ as the confidence level $f$ is given $x$. Accordingly, data with low confidence are supposed to be useful to uncover faults. 

\textbf{TestRank}~\cite{NEURIPS2021_ae785101} is a learning-based method that contains three main steps. First, it extracts two types of features from the input data, 1) the output from the logits layer as intrinsic attributes, 2) the graph information that contains the distance of this data (cosine distance) to others, and the label of this data.  Then, a GNN model is built to learn the graph information and predict the learned contextual attributes of the data. Finally, the contextual attributes and intrinsic attributes are combined and fed to a simple binary classification model to learn the failure-revealing capability.

\textbf{PRIMA}~\cite{9402064} is another learning-based method that first generates mutants from both input data and models. Then, it extracts features of each input data from those mutants, e.g., the number of mutants that have different predicted labels from the original model on this data. Finally, PRIMA trains a ranking model (XGBoost ranking algorithm) based on the features of correctly and wrongly predicted data to identify if the unlabeled data is a fault.

\textbf{Adaptive test selection (ATS)}~\cite{9793868} is a method that totally depends on the final output probability of the data. Roughly speaking, it first projects the output vectors (built by the top-3 maximum vectors) to a plane and then calculates the coverage of each data on the plane. After that, the difference between the coverage of a single data and the whole test set is utilized as an identifier to distinguish the faults and normal test data. 

\subsection{Performance Estimation Methods}

Given a set of unlabeled test data and a DNN model, performance estimation methods tend to select a representative subset from the entire test data. The model performance on this subset should be close to that on the entire test data. In this way, developers can quickly observe how the model performs on unseen data and then take the next step. For example, if the performance is lower than the exception, developers need to repair the model.

\textbf{Cross Entropy-based Sampling (CES)}~\cite{li2019boosting} selects a subset of test data that have the minimum cross entropy with the entire set.

\textbf{Practical accuracy estimation (PACE)}~\cite{chen2020practical} first clusters data into groups based on the hierarchical density-based spatial clustering of applications with noise algorithm. Next, from each group, PACE proportionally selects data to represent the entire group.

Note that, prediction outputs from intermediate layers of DNNs are required to calculate the cross entropy in CES and to conduct clustering in PACE. 

\subsection{Pitfalls of Test Selection Methods}
\label{subsec:weakness_analysis}
We conjecture that there are three types of pitfalls that hinder the success of test selection methods.

\textbf{Blind in high uncertainty but correctly predicted data.} All uncertainty-based fault detection methods (DeepGini, MCP, MC, MaxP, PRIMA, and ATS) assume that data with high uncertainty are more likely to be misclassified by the model. However, this assumption only stands when the model is well-trained and has low bias. DSA and TestRank do not rely on this assumption. They utilize the \textit{difference} between correctly predicted data and wrongly predicted data to determine the new faults. However, in the original test data, there are not many high-uncertainty but correctly predicted faults which makes the learned \textit{difference} can not be generalized to this type of fault. We conjecture that all the fault detection methods will identify the high uncertainty but correctly predicted data as faults. And in this paper, we call this type of data as \textit{\textbf{Type1}} data. The methods involved are DSA, DeepGini, MCP, MC, MaxP, TestRank, PRIMA, and ATS.

\textbf{Blind in low uncertainty but wrongly predicted data.} The same reason as the last point, we conjecture that all the fault detection methods cannot detect the low uncertainty but wrongly predicted fault data, and we call this type of data as \textit{\textbf{Type2}} data. The methods involved are DSA, DeepGini, MCP, MC, MaxP, TestRank, PRIMA, and ATS.

\textbf{Susceptible to layer selection.} We found that all the existing model performance estimation methods rely on the intermediate output of the model. However, a DNN model normally consists of multiple hidden layers. We doubt that the effectiveness of such test selection methods is highly impacted by the choice of outputs of hidden layers, and it is difficult to make a clear conclusion about which layer is better across different datasets and models. The methods involved are DSA, CES, and PACE.

\section{Overview}
\label{sec:empirical}

In this section, we design a study to investigate if the potential non-robust features exist. Although a perfect test selection method may not exist, people should know what they need to pay attention to when proposing, evaluating, and applying test selection methods.

\subsection{Study Design}

\begin{figure*}[ht]
	\centering
	\includegraphics[width=\textwidth]{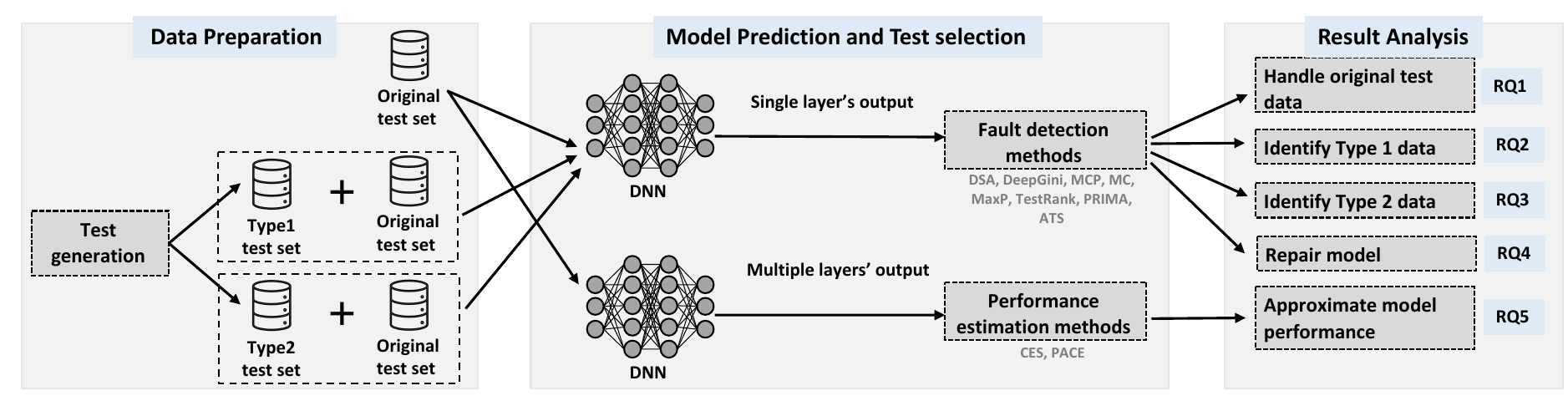}
	\caption{Overview of the study design.
	}
	\label{fig:overview}
\end{figure*}

Figure~\ref{fig:overview} illustrates the overview workflow of our empirical study. Overall, our plan is to explore and answer the following five research questions:

\noindent\textbf{RQ1: }\textit{How do fault detection methods compare to each other?}

\noindent\textbf{RQ2: }\textit{Can existing fault detection methods bypass uncertain but correctly classified data?}

\noindent\textbf{RQ3: }\textit{Can fault detection methods identify high confidence but wrongly classified faults?}

\noindent\textbf{RQ4: }\textit{How do uncertain (high confidence) but correctly (wrongly) classified data affect test selection-based model repair?}

\noindent\textbf{RQ5: }\textit{How does the choice of intermediate layers affect performance estimation methods?}

Before studying the problems listed in Sec~\ref{subsec:weakness_analysis}, we first investigate the common concern of using test selection methods -- do we have the best choice among the massive number of methods? To do this, we have a quick look at all fault detection methods and check, when we only consider the original test data, if there is a recommended method that performs consistently better than others, or if there are methods that have big performance variances across different datasets that are not recommended to use.

Then, we go deeper to study the robustness of test selection methods. For fault detection, as discussed in Section~\ref{subsec:weakness_analysis}, there are two types of test data that are difficult for existing test selection methods to handle, \textit{Type1}: high uncertainty but correctly predicted data, and \textit{Type2}: low uncertainty but wrongly predicted data. The first step of our study is to generate these two types of test data. We follow the previous work~\cite{zhang2020towards} and design a genetic algorithm (GA) based test generation technique. The details will be introduced in the next section. After the data preparation, we inject the generated test data into the original test data and obtain three groups of test data, 1) only original test data, 2) \textit{Type1} test data + original test data, and 3) \textit{Type2} test data + original test data. Then, we perform test selection on these three groups of test data and analyze the effectiveness of each method. In addition, since the final target of fault detection is to fix or repair the pre-trained model to make it bypass these faults, we conduct a study to utilize the selected test data to retrain (the common model repair approach) the model. Specifically, we first evenly split the three groups of test data into two parts, the candidate set and the new test set. Then, we perform test selection on the candidate set and combine the selected data with training data to retain the model with a few epochs. After that, we test the retrained model using the new test set and analyze the effectiveness of test selection methods when facing \textit{Type1} and \textit{Type2} test data in terms of model repair. 

On the other hand, for the test selection-based model performance estimation, we choose outputs from different hidden layers gained by using the original test data and then fed them to the test selection methods. We analyze the difference in the estimated model performance to check the impact of the choice of the intermediate outputs. 

\subsection{Test Data Preparation}

Test generation is a common practice to prepare test data when testing DNNs. In which, genetic algorithm~\cite{zhang2020towards,gao2020fuzz} based test generation technique is effective in generating diverse test data. To prepare the two types of test data, we design a very simple and flexible (easy to extend) GA-based test generation technique, as presented in Algorithm~\ref{alg:tg}. Mention that, our purpose is to reveal the limitations of test selection methods instead of attacking them. Other methods like modified white or black-box adversarial attacks can also generate such kinds of test data. Investigation of them is left as future work. 

\begin{algorithm}[htpb]
\small
\caption{GA-based test generation}
\label{alg:tg}
\SetAlgoLined
\Input{$seed$: seed data\\
$pop\_size$: size of population\\
$max\_iteration$: maximum number of iteration\\
$tour\_size$: size of tour data\\
}
\Output{$X\_generated$: generated test data}

$pop = Population\_Initialization(seed,\ pop\_size)$ \\

$count\_num = 0$ \\

\While{$count\_num < max\_iteration$}
{
    \For{$X\_generated\ in\ pop$}
    {
        \If {$X\_generated$ fits $Condition$}
        {
            \Return $X\_generated$;
        }
    }
    $new\_pop\ = \ []$ \\
    $fitness = Fitness\_Calculation()$ \\
    $individual\ =\ Select\_Best(pop,\ fitness)$\\
    $pop\ =\ pop\setminus individual$ \\
    $new\_pop.update(individual)$\\
    $pop\ =\ Crossover(pop,\ tour\_size)$ \\
    $pop\ =\ Mutation(pop)$ \\
    $new\_pop.update(pop)$ \\
    $pop\ =\ new\_pop$ \\
}
$X\_generated\ =\ pop[0]$ \\

\Return $X\_generated$;
\end{algorithm}

Here, we explain the flexible components in the generation process that control whether to generate \textit{Type1} or \textit{Type2} test data. We use the notations defined in Section~\ref{sec:mertics}. Given a DNN model $f$, a test data $x$ and its true label $y$, $p_{\widehat{y}}\left(x\right)$ represents the probability of $x$ belonging to the predicted class $\widehat{y}$ by $f$. $y_s$ is the predicted second most likely class.

\textbf{Seed preparation.} We follow a previous work~\cite{xie2019deephunter} to randomly select seed data from each class of the datasets to increase the diversity of the seeds.

\textbf{Condition} controls if \textit{Type1} or \textit{Type2} test data are successfully generated. For \textit{Type1} test data, the generated data should fit all these three conditions: 1) $\widehat{y}= y$, 2) $p_{\widehat{y}}\left(x\right)\leq T_1$, and 3) $p_{\widehat{y}}\left(x\right)-p_{y_s}\left(x\right)\leq T_2$. For \textit{Type2} test data, the generated data should fit all these two conditions: 1) $\widehat{y}\neq y$, 2) $p_{\widehat{y}}\left(x\right)>T_1$. Here, $T_1$ and $T_2$ are two predefined thresholds (see Table~\ref{tab:configuration}). 
\textbf{Fitness function} controls the evolution direction during test generation. For \textit{Type1} data generation, the fitness function is the combination of $-\left(p_{\widehat{y}}\left(x\right)-p_{y_s}\left(x\right)\right)$ and $-p_{\widehat{y}}\left(x\right)$. For \textit{Type2}, the fitness function is $p_{y_s}\left(x\right)$ if the data are correctly classified, otherwise $p_{\widehat{y}}\left(x\right)$.

\textbf{Crossover} follows the work~\cite{zhang2020towards} and utilizes the tournament selection strategy to select two tournaments, and then chooses one data with the biggest fitness score from each tournament respectively to do randomly pixel changing.

\textbf{Mutation} aims to increase the diversity of the population. Here, we utilize image transformation techniques to generate mutants and control the perturbation size to ensure the semantics of the generated image do not change. 

\subsection{Experimental Setup}
\label{sec:setup}

\textbf{Dataset and model.} Table~\ref{tab:dataandmodel} lists the details of our studied datasets and models. MNIST~\cite{lecun1998gradient} and SVHN~\cite{svhn2011} contain digital numbers. CIFAR10~\cite{Alex2009techm} is a collection of color images with 10 categories (e.g., airplane, bird). CIFAR100~\cite{lecun1998gradient} is a more challenging version of CIFAR10 with fine-grained 100 categories (e.g., aquarium fish, flatfish). Traffic-Sign~\cite{Houben-IJCNN-2013} contains traffic sign images and is commonly used for self-driving cars. For each dataset, we build two popular models that are mainly from the LeNet~\cite{lecun2015lenet}, ResNet~\cite{7780459}, VGG~\cite{simonyan2014very}, and DenseNet~\cite{huang2017densely} families.  

\begin{table}[h]
\centering
\caption{Dataset and Model.}
\label{tab:dataandmodel}
\resizebox{.9\columnwidth}{!}{
\begin{tabular}{lcllc}
\hline
\textbf{Dataset} & \multicolumn{1}{l}{\textbf{Class Number}} & \textbf{Test Size} & \textbf{Model} & \multicolumn{1}{l}{\textbf{Test Accuracy (\%)}} \\ \hline
\multirow{2}{*}{MNIST} & \multirow{2}{*}{10} & \multirow{2}{*}{10000} & LeNet1 & 98.07 \\
 &  &  & LeNet5 & 98.87 \\ \hline
\multirow{2}{*}{SVHN} & \multirow{2}{*}{10} & \multirow{2}{*}{10000} & LeNet5 & 87.55 \\
 &  &  & ResNet20 & 95.85 \\ \hline
\multirow{2}{*}{CIFAR10} & \multirow{2}{*}{10} & \multirow{2}{*}{10000} & ResNet20 & 87.44 \\
 &  &  & VGG16 & 91.39 \\ \hline
\multirow{2}{*}{Traffic-Sign} & \multirow{2}{*}{43} & \multirow{2}{*}{12630} & LeNet5 & 83.37 \\
 &  &  & VGG16 & 93.08 \\ \hline
\multirow{2}{*}{CIFAR100} & \multirow{2}{*}{100} & \multirow{2}{*}{10000} & ResNet50 & 76.89 \\
 &  &  & DenseNet121 & 71.72 \\ \hline
\end{tabular}
}
\end{table}

\textbf{Configurations of test generation.} There are some hyper-parameters in the test generation process. Investigating the best configuration is not our focus. Instead, we give recommended configurations used in our study that can already achieve our purpose. Table~\ref{tab:configuration} lists the detailed configurations. And for the image mutation, we employ four image transformation techniques, image contrast changing, image brightness changing, blur noise adding, and Gaussian noise adding~\cite{hendrycks2019benchmarking}. It is flexible and easy to extend with more mutation operators. To preserve the semantics of the generated data, we follow the work~\cite{xie2019diffchaser} and limit the maximum L-infinite perturbation size of the injected noise. 

\begin{table}[h]
\centering
\caption{Configurations of test generation.}
\label{tab:configuration}
\resizebox{.9\columnwidth}{!}{
\begin{tabular}{lccccccc}
\hline
\multirow{2}{*}{} & \multicolumn{2}{c}{\textbf{Type1}} & \textbf{Type2} & \multirow{2}{*}{\begin{tabular}[c]{@{}c@{}}\textbf{perturbation}\\ \textbf{size}\end{tabular}} & \multirow{2}{*}{\textbf{\begin{tabular}[c]{@{}c@{}}pop\\ size\end{tabular}}} & \multirow{2}{*}{\textbf{\begin{tabular}[c]{@{}c@{}}max\\ iteration\end{tabular}}} & \multirow{2}{*}{\textbf{\begin{tabular}[c]{@{}c@{}}tour\\ size\end{tabular}}} \\
 & \textbf{T\_1} & \textbf{T\_2} & \textbf{T\_1} &  &  &  &  \\ \hline
\textbf{MNIST} & 0.5 & 0.01 & 0.95 & 0.5 & \multirow{4}{*}{200} & \multirow{4}{*}{200} & \multirow{4}{*}{20} \\
\textbf{SVHN}, \textbf{CIFAR10} & 0.5 & 0.01 & 0.95 & 0.05 &  &  &  \\
\textbf{Traffic} & 0.3 & 0.05 & 0.95 & 0.05 &  &  &  \\
\textbf{CIFAR100} & 0.1 & 0.05 & 0.5 & 0.05 &  &  &  \\ \hline
\end{tabular}
}
\end{table}

\textbf{Configurations of test selection.} For the fault detection, we set the maximum labeling budget as 10\% of the entire test data, which is a common setting in previous works~\cite{9793868,hu2022empirical,9286133}. For the performance estimation, we follow previous works~\cite{li2019boosting,chen2020practical} and set the labeling budgets from 50 to 180 in intervals of 10. For the intermediate output selection for the performance estimation, we chose the last 1, 2, and 3 hidden layers in our study.

\textbf{Evaluation methods.}
To evaluate the effectiveness of fault detection, we adopt the measurement, Test Relative Coverage (TRC), from the literature~\cite{NEURIPS2021_ae785101}. 

TRC is defined as:

\begin{equation}
    TRC\left(X\right)=\frac{|F\_detected|}{\min\left(|F\_X|, Budgets\right)}
\end{equation}
where $Budgets$ is the size of selected data. 

Besides, we employ Student's $t$-test~\cite{owen1965power} which is a famous statistical analysis method to analyze the significance of the impact from the \textit{Type1} and \textit{Type2} data. For model repair, we use the absolute accuracy difference between the original model and the repaired model to quantify the effectiveness of related test selection methods.

\textbf{Implementation.} The main framework uses TensorFlow~\cite{abadi2016tensorflow} 2.3.0. The implementation of each test selection method is modified from the source code provided by the original paper to fit our experiment environment. All experiments run on a 2.6 GHz Intel Xeon Gold 6132 CPU with an NVIDIA Tesla V100 16G SXM2 GPU. We repeat the experiments with randomness factors five times and report the average results, e.g., test selection using PRIMA, the model repair process.

\section{Results}
\label{sec:results}

\subsection{RQ1: Performance on Original Test Data}

\begin{figure*}[ht]
    \centering
    \subfigure[MNIST, LeNet-1]{\label{subfig:lenet1}
    \includegraphics[scale=0.21]{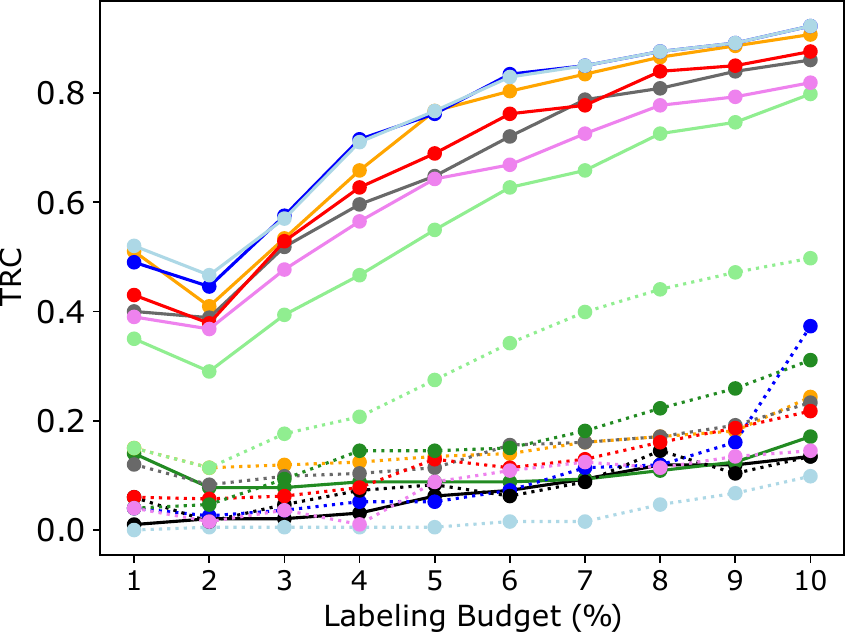}%
    }
    \subfigure[MNIST, LeNet5]{\label{subfig:lenet5}
    \includegraphics[scale=0.21]{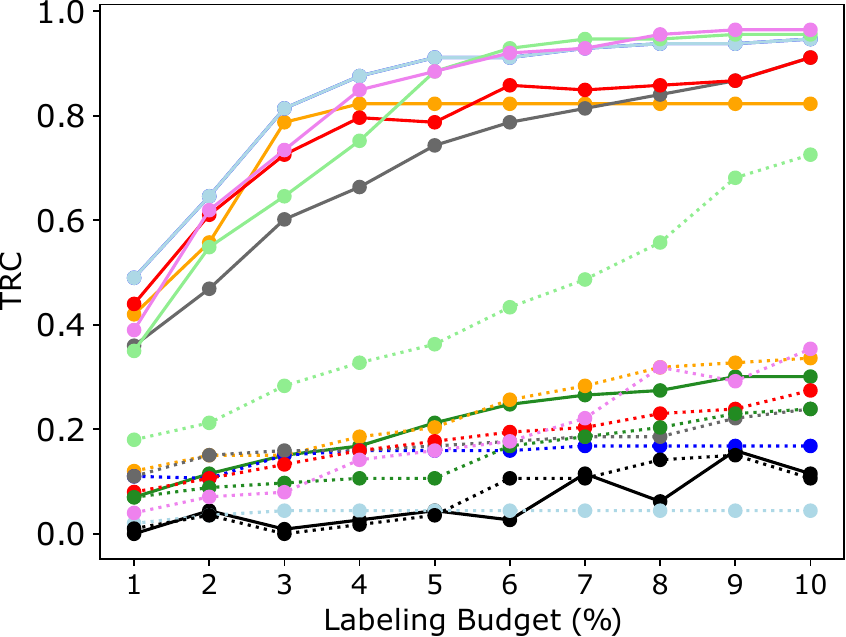}%
    }
    \subfigure[SVHN, LeNet-5]{
    \includegraphics[scale=0.21]{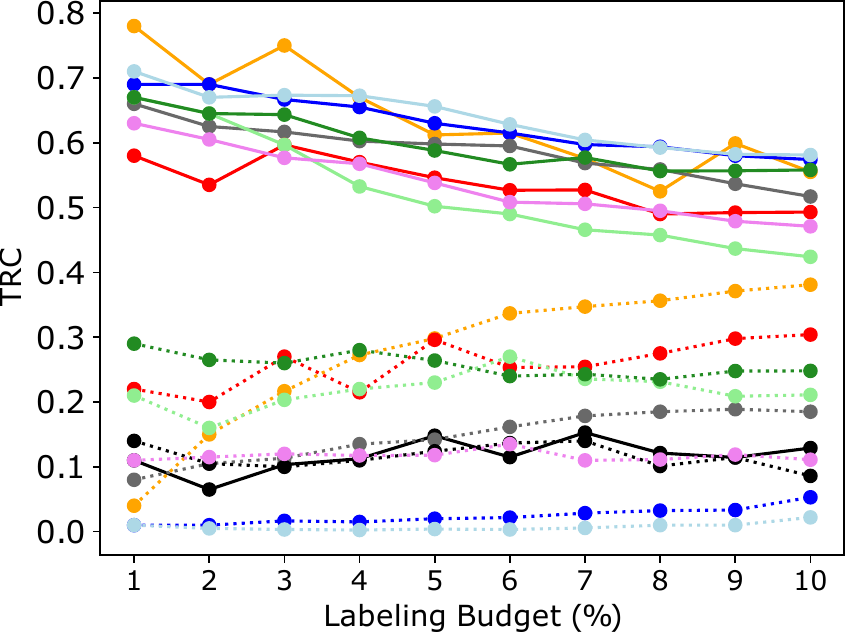}%
    }
    \subfigure[SVHN, ResNet20]{
    \includegraphics[scale=0.21]{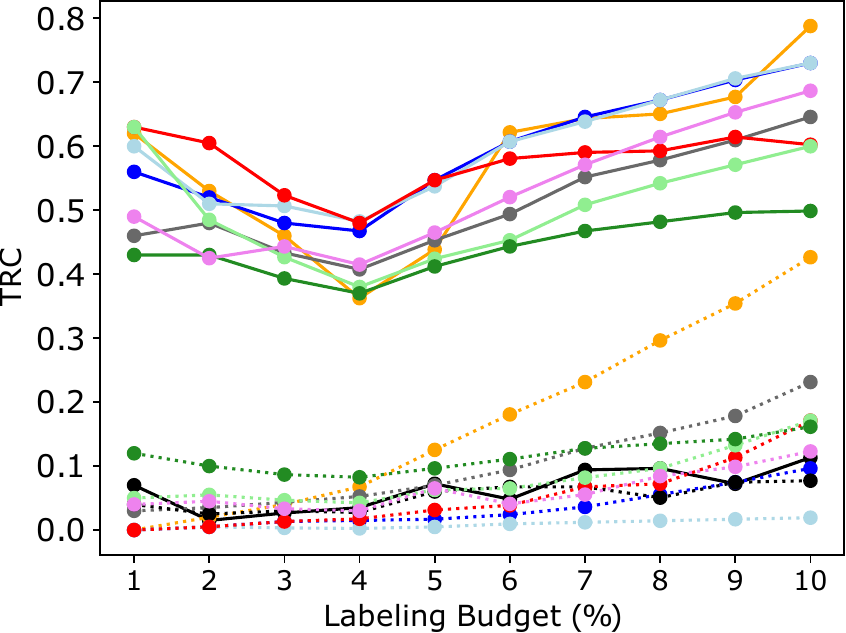}%
    }
    \subfigure[CIFAR10, ResNet20]{
    \includegraphics[scale=0.21]{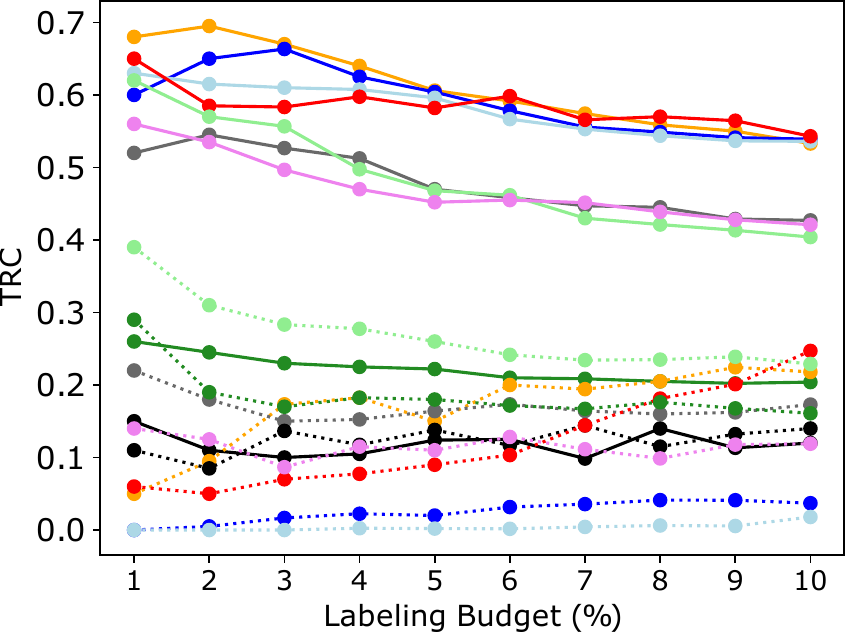}
    }
    \subfigure[CIFAR10, VGG16]{
    \includegraphics[scale=0.21]{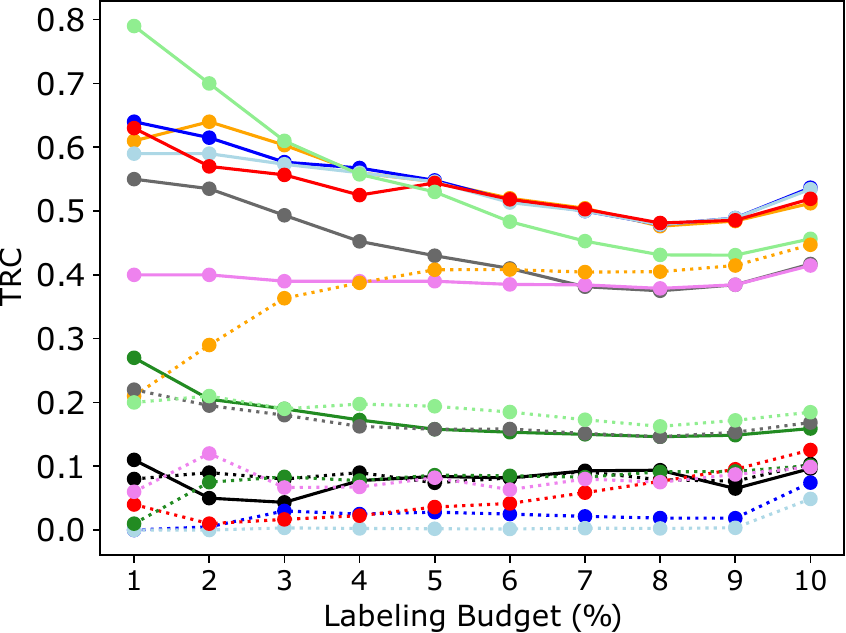}%
    }
    \subfigure[Traffic, LeNet5]{
    \includegraphics[scale=0.21]{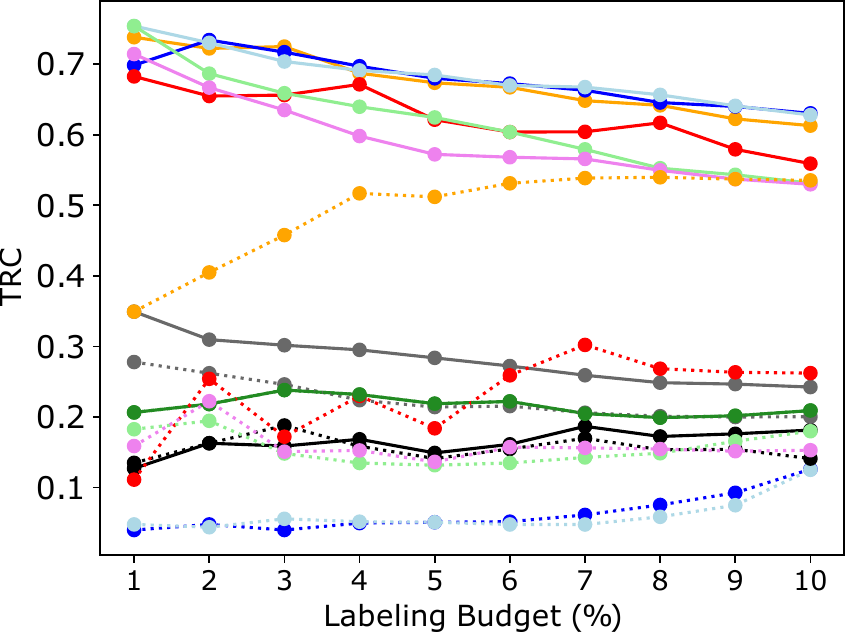}%
    }
    \subfigure[Traffic, VGG16]{
    \includegraphics[scale=0.21]{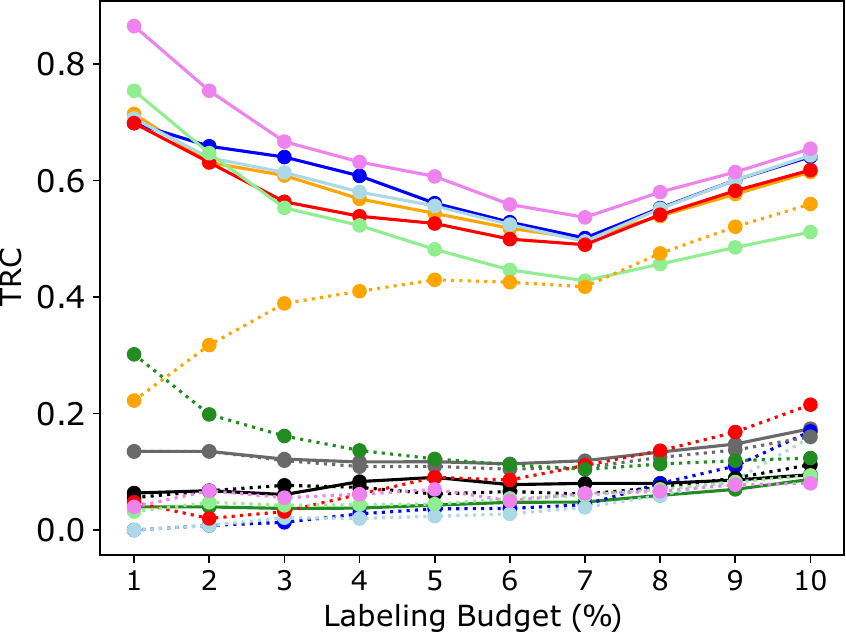}%
    }
    \subfigure[CIFAR100, ResNet50 ]{
    \includegraphics[scale=0.21]{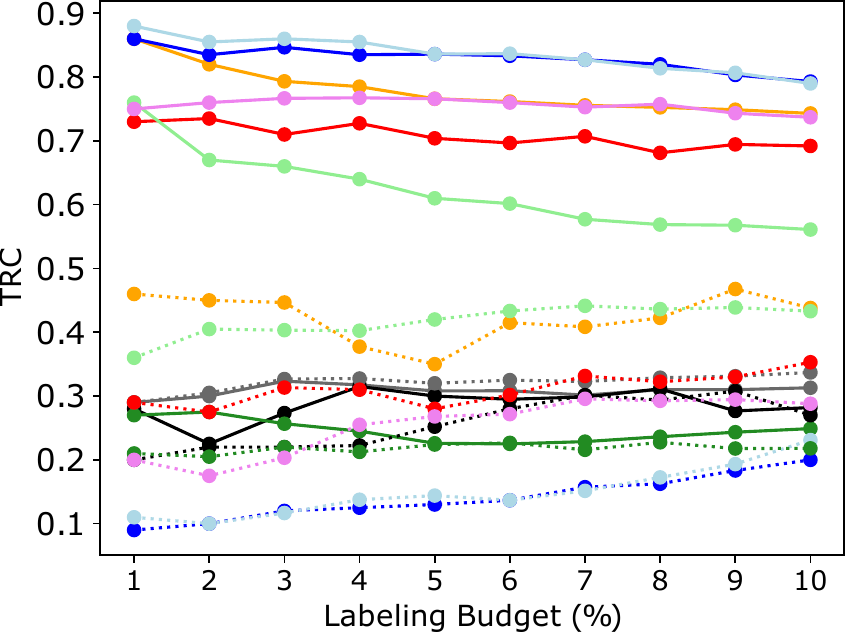}%
    }
    \subfigure[CIFAR100, DenseNet121]{
    \includegraphics[scale=0.21]{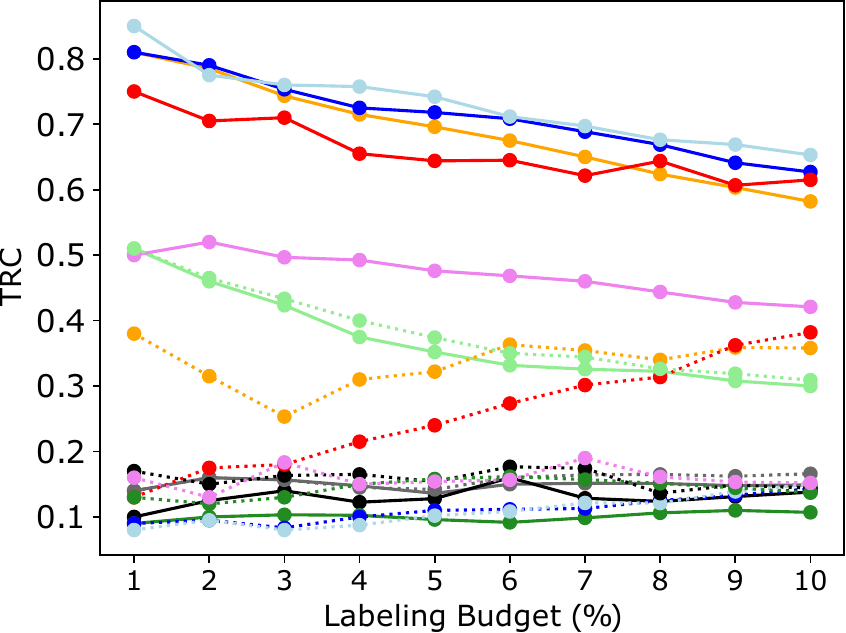}
    }
    \subfigure{\label{subfig:legend}
    \includegraphics[scale=0.23]{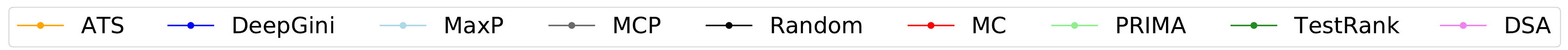}%
    }
    \caption{Test Relative Coverage~(TRC) of Test Selection Methods on Original and \textit{Type1} Test Data. Solid lines: original test data, dash lines: \textit{Type1} test data. }
    \label{fig:rq1}
\end{figure*}

\begin{wrapfigure}[12]{c}{0.55\linewidth}
 \centering
 \includegraphics[width =\linewidth]{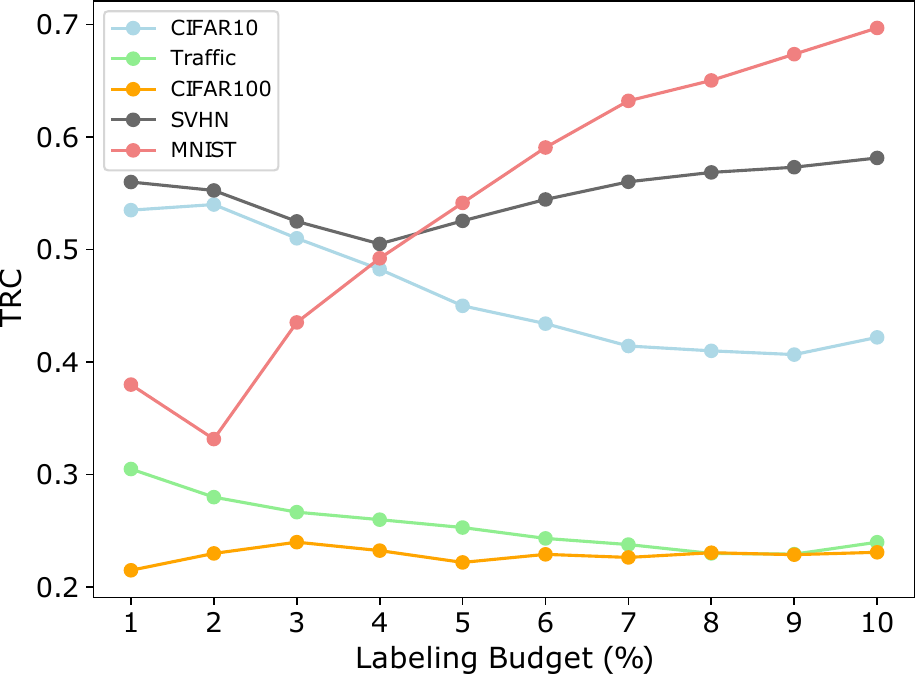}
 \caption{TRC of MCP on different datasets.}\label{fig:rq1_mcp_analysis}
\end{wrapfigure}

First, for the practical usage of test selection methods, we need to know if there is one that outperforms others and can be recommended to use. Figure~\ref{fig:rq1} depicts the test relative coverage of each method. From the results, we can see the TRC of each method variant from different datasets and models, which means that when facing new datasets, it is hard to choose which one we should use.  More specifically, Table~\ref{tab:classnumber} shows the TRC of each method when the labeling budget is 10\%. From the ranking of each method, we can draw the same conclusion as before that no method can always stand out. However, we found two methods, DeepGini and MaxP, have top-3 TRC scores in all situations. This finding is similar to the recent study~\cite{weiss2022simple} which reveals that simple methods work better for DL-targeted test selection. 

\begin{table}[ht]
\centering
\caption{Fault detection performance of each method with labeling budget of 10\% -- TRC(ranking).
}
\label{tab:classnumber}
\resizebox{\columnwidth}{!}
{
\begin{tabular}{lccccccccc}
\hline
 & \textbf{Random} & \textbf{DSA} & \textbf{DeepGini} & \textbf{MCP} & \textbf{MC} & \textbf{MaxP} & \textbf{TestRank} & \textbf{PRIMA} & \textbf{ATS} \\ \hline
\textbf{MNIST-LeNet1} & 0.13(9) & 0.82(6) & 0.92(1) & 0.86(5) & 0.88(4) & 0.92(1) & 0.17(8) & 0.80(7) & 0.91(3) \\
\textbf{MNIST-LeNet5} & 0.12(9) & 0.96(1) & 0.95(3) & 0.91(5) & 0.91(5) & 0.95(3) & 0.30(8) & 0.96(1) & 0.82(7) \\
\textbf{SVHN-LeNet5} & 0.13(9) & 0.47(7) & 0.57(2) & 0.52(5) & 0.49(6) & 0.58(1) & 0.56(3) & 0.42(8) & 0.56(3) \\
\textbf{SVHN-ResNet20} & 0.11(9) & 0.69(4) & 0.73(2) & 0.65(5) & 0.60(6) & 0.73(2) & 0.50(8) & 0.60(6) & 0.79(1) \\
\textbf{CIFAR10-ResNet20} & 0.12(9) & 0.42(6) & 0.54(1) & 0.43(5) & 0.54(1) & 0.54(1) & 0.20(8) & 0.40(7) & 0.53(4) \\
\textbf{CIFAR10-VGG16} & 0.10(9) & 0.41(7) & 0.54(1) & 0.42(6) & 0.52(3) & 0.53(2) & 0.16(8) & 0.46(5) & 0.51(4) \\
\textbf{Traffic-LeNet5} & 0.18(9) & 0.53(5) & 0.63(1) & 0.24(7) & 0.56(4) & 0.63(1) & 0.21(8) & 0.53(5) & 0.61(3) \\
\textbf{Traffic-VGG16} & 0.09(8) & 0.65(1) & 0.64(2) & 0.17(7) & 0.62(4) & 0.64(2) & 0.09(8) & 0.51(6) & 0.61(5) \\
\textbf{CIFAR100-ResNet50} & 0.28(8) & 0.74(3) & 0.79(1) & 0.31(7) & 0.69(5) & 0.79(1) & 0.25(9) & 0.56(6) & 0.74(3) \\
\textbf{CIFAR100-DenseNet101} & 0.14(8) & 0.42(5) & 0.63(2) & 0.15(7) & 0.62(3) & 0.65(1) & 0.11(9) & 0.30(6) & 0.58(4) \\ \hline
\textbf{Average} & 0.14(9) & 0.61(5) & 0.69(2) & 0.47(7) & 0.64(4) & 0.70(1) & 0.25(8) & 0.55(6) & 0.67(3) \\
\textbf{Variance} & 0.00(1) & 0.04(7) & 0.02(2) & \textbf{0.07(9)} & 0.02(2) & 0.02(2) & 0.02(2) & 0.04(7) & 0.02(2) \\ \hline
\end{tabular}
}
\end{table}

Interestingly, besides proving the literature findings, we found that the TRC of MCP has a bigger variance across different datasets than others. When checking the TRC of MCP of each dataset, we can see MCP has the same level of performance as DeepGini and MaxP~(e.g., 0.91 vs. 0.95 vs. 0.95) in datasets MNIST, SVHN, and CIFAR10. But when checking the results on Traffic and CIFAR100, the performance of MCP has a great degradation, and the difference between MCP, DeepGini, and MaxP becomes not negligible~(e.g., 0.15 vs. 0.63 vs. 0.65). The reason for this unstable performance of MCP is that MCP is highly dependent on the number of classes of the dataset. MCP first divides data into fine-grained boundary areas before selecting. For a 10-class dataset (e.g., MNIST), the number of decision boundary areas is $A_{10}^{2}=90$. However, when the class number increases to 100 (e.g., CIFAR100), the number of boundary areas also increases to $A_{100}^{2}=9900$, which can easily exceed the labeling budget. In this case, MCP becomes a random-like selection method. Figure~\ref{fig:rq1_mcp_analysis}, the detailed TRC scores of MCP on different datasets, also confirm our analysis. MCP performs much worse on Traffic and CIFAR100 than on others.

Then, we go further and check the efficiency of each method. Table~\ref{tab:time_analysis} presents the executing time of each test selection method when it ranks all the test data once. It is reasonable that the time cost increases along with the increasing complexity of datasets and model architectures. However, we found that, in the case of (SVHN-LeNet5 vs Traffic-LeNet5) and (CIFAR10-VGG16 vs Traffic-VGG16), only ATS and PRIMA have a big difference in the time cost. PRIMA has complex steps including preparing the mutants of data and models, and the total time cost is unstable and hard to analyze. For the ATS, the time cost increases significantly because 1) it tries to identify the faults from each class one by one, and 2) when the number of classes increases, the types of fault pattern (combination of classes) increase non-linearly, e.g., from $A_{10}^{3}$ to $A_{100}^{3}$. 

\begin{table}[ht]
\centering
\caption{Running time (seconds) of fault detection target test selection methods when ranking all the test data once.}
\label{tab:time_analysis}
\resizebox{\columnwidth}{!}
{
\begin{tabular}{lccccccccc}
\hline
 & \textbf{Random} & \textbf{DSA} & \textbf{DeepGini} & \textbf{MCP} & \textbf{MC} & \textbf{MaxP} & \textbf{TestRank} & \textbf{PRIMA} & \textbf{ATS} \\ \hline
\textbf{MNIST-LeNet1} & 0.44 & 10.26 & 11.50 & 0.61 & 1.11 & 0.45 & 30.90 & 185.38 & 18.92 \\
\textbf{MNIST-LeNet5} & 0.48 & 11.98 & 12.58 & 0.64 & 1.20 & 0.49 & 32.21 & 289.33 & 3.17 \\
\textbf{SVHN-LeNet5} & 0.77 & 12.06 & 18.57 & 0.93 & 1.74 & 0.79 & 27.86 & 286.37 & 2.06 \\
\textbf{SVHN-ResNet20} & 1.56 & 51.92 & 39.66 & 1.81 & 3.14 & 1.53 & 43.38 & 2085.41 & 19.08 \\
\textbf{CIFAR10-ResNet20} & 1.59 & 51.02 & 40.38 & 1.77 & 3.35 & 1.57 & 54.38 & 1980.69 & 31.24 \\
\textbf{CIFAR10-VGG16} & 1.17 & 131.21 & 27.06 & 1.31 & 2.54 & 1.19 & 76.70 & 3986.26 & 18.34 \\
\textbf{Traffic-LeNet5} & 0.93 & 14.09 & 22.24 & 1.08 & 2.14 & 0.96 & 35.94 & 2568.60 & 1035.86 \\
\textbf{Traffic-VGG16} & 1.37 & 101.23 & 31.44 & 1.52 & 3.01 & 1.38 & 76.40 & 2658.98 & 1475.31 \\
\textbf{CIFAR100-renset50} & 2.66 & 62.42 & 58.82 & 2.85 & 5.53 & 2.66 & 60.83 & 3183.44 & 5080.69 \\
\textbf{CIFAR100-denset101} & 4.82 & 58.49 & 102.55 & 5.00 & 9.76 & 4.76 & 56.70 & 3042.73 & 4170.33 \\ \hline
\end{tabular}
}
\end{table}

\noindent
{\framebox{\parbox{0.96\linewidth}{
\textbf{Answer to RQ1}: No methods have consistently better performance than others. MCP and ATS have significant effectiveness and efficiency drops, respectively, when handling data with a large number of classes.}}}

\subsection{RQ2: Impact of \textit{Type1} Test Data}

To study how the test selection methods deal with the data that are correctly classified but with high prediction uncertainty. We generate \textit{Type1} test data and inject it into the original test set, then evaluate the test selection methods accordingly. Here, the number of injected data is the same as the budget of selected (labeled) data. 

Figure~\ref{fig:rq1} presents the results of the 10 fault detection methods with the labeling budget ranging from 1\% to 10\%. In general, there is a clear gap between the results of the original test data and the \textit{Type1} test data. All the test selection methods perform worse when the \textit{Type1} data exists, which means that methods tend to select \textit{Type1} data as the faults but in fact, they are not. Then, comparing each method, PRIMA and ATS are more effective than others when dealing with \textit{Type1} test data, e.g., in MNIST, PRIMA is significantly better than other methods, and in SVHN, Traffic, and CIFAR100 datasets, ATS is better. The potential reason is that PRIMA mutates the data multiple times and then calculates the uncertainty scores. Although the generated test data are close to the decision boundary, their predictions may not change after mutation, thus, PRIMA does not identify these \textit{Type1} test data as faults. ATS selects faults from diverse fault patterns and does not only select uncertain data. Besides, we statically compare the fault detection performance of each method on original test data and \textit{Type1} data using $t$-test. The results show that except for \textit{Random}, \textit{TestRank}, and \textit{MCP}, all the methods perform significantly worse on \textit{Type1} data than on original test data (with a p-value~\textless 0.05).

\begin{table*}[]
\centering
\caption{Average TRC values of test selection methods over all datasets and models (Labeling Budget 10\%). }
\label{tab:type1_test_data}
\resizebox{.8\textwidth}{!}
{
\begin{tabular}{ccccccccccc}
\hline
\multicolumn{1}{l}{} & \textbf{Random} & \textbf{DSA} & \textbf{DeepGini} & \textbf{MCP} & \textbf{MC} & \textbf{MaxP} & \textbf{TestRank} & \textbf{PRIMA} & \textbf{ATS} & \multicolumn{1}{l}{\textbf{Average}} \\ \hline
\textbf{Ori} & 0.14 & 0.61 & 0.69 & 0.47 & 0.64 & 0.70 & 0.25 & 0.55 & 0.67 & 0.53 \\
\textbf{Type1} & 0.13 & 0.16 & 0.14 & 0.21 & 0.26 & 0.09 & 0.19 & 0.30 & 0.39 & 0.21 \\
\textbf{Diff} & 6.42\% \contour{red}{\textcolor{red}{$\downarrow$}} & 73.47\% \contour{red}{\textcolor{red}{$\downarrow$}} & 79.25\% \contour{red}{\textcolor{red}{$\downarrow$}} & 55.05\% \contour{red}{\textcolor{red}{$\downarrow$}} & 60.31\% \contour{red}{\textcolor{red}{$\downarrow$}} & 86.85\% \contour{red}{\textcolor{red}{$\downarrow$}} & 24.96\% \contour{red}{\textcolor{red}{$\downarrow$}} & 45.25\% \contour{red}{\textcolor{red}{$\downarrow$}} & 40.88\% \contour{red}{\textcolor{red}{$\downarrow$}} & 52.49\% \contour{red}{\textcolor{red}{$\downarrow$}} \\ \hline
\end{tabular}
}
\vspace{-5mm}
\end{table*}

More specifically, Table~\ref{tab:type1_test_data} presents the average test relative coverage values of each test selection method among all the datasets and models when the labeling budget is 10\%. We can see that except for random selection, only TestRank has a small performance degradation on the \textit{Type1} test data, but its TRC on the original test data is not high. Other test selection methods have at least 42\% test relative coverage drops, where MaxP and DeepGini drop the greatest.

\noindent \\
{\framebox{\parbox{0.96\linewidth}{
\textbf{Answer to RQ2}: Existing fault detection methods cannot distinguish real faults and \textit{Type1}~(correctly predicted but uncertain) test data. On average, when facing \textit{Type1} test data, the test relative coverage of test selection methods drops 52.99\%.}}}

\subsection{RQ3: Impact of \textit{Type2} Test Data}

Next, we explore how the data that are wrongly classified but with high prediction confidence affect the effectiveness of fault detection methods. Same as the last study, we inject the same number of labeling budget of \textit{Type2} test data into the original test data and then perform the fault detection. 

\begin{table*}[h]
\centering
\caption{Results of TRC on \textit{Type2} test data (labeling budget 10\%).}
\label{tab:type2_test_data}
\resizebox{1.7\columnwidth}{!}
{
\begin{tabular}{llccccccccccc}
\hline
\multicolumn{3}{l}{} & \textbf{Random} & \textbf{DSA} & \textbf{MC} & \textbf{DeepGini} & \textbf{MCP} & \textbf{MaxP} & \textbf{TestRank} & \textbf{PRIMA} & \textbf{ATS} & \multicolumn{1}{l}{\textbf{Average}} \\ \hline
\multirow{4}{*}{\textbf{MNSIT}} & \multirow{2}{*}{\textbf{LeNet1}} & \textbf{All} & 0.11 & 0.13 & 0.18 & 0.18 & 0.24 & 0.18 & 0.24 & 0.14 & 0.18 & 0.17 \\
 &  & \textbf{Type2 only} & 0.08 & 0.11 & 0.00 & 0.00 & 0.07 & 0.00 & 0.20 & 0.04 & 0.00 & 0.06 \\
 & \multirow{2}{*}{\textbf{LeNet5}} & \textbf{All} & 0.10 & 0.10 & 0.10 & 0.11 & 0.20 & 0.11 & 0.16 & 0.12 & 0.15 & 0.13 \\
 &  & \textbf{Type2 only} & 0.09 & 0.09 & 0.00 & 0.00 & 0.10 & 0.00 & 0.12 & 0.06 & 0.06 & 0.06 \\ \hline
\multirow{4}{*}{\textbf{SVHN}} & \multirow{2}{*}{\textbf{LeNet5}} & \textbf{All} & 0.22 & 0.21 & 0.50 & 0.57 & 0.53 & 0.58 & 0.46 & 0.45 & 0.57 & 0.45 \\
 &  & \textbf{Type2 only} & 0.11 & 0.09 & 0.04 & 0.00 & 0.01 & 0.00 & 0.00 & 0.08 & 0.00 & 0.04 \\
 & \multirow{2}{*}{\textbf{ResNet20}} & \textbf{All} & 0.14 & 0.13 & 0.26 & 0.30 & 0.30 & 0.30 & 0.90 & 0.25 & 0.30 & 0.32 \\
 &  & \textbf{Type2 only} & 0.11 & 0.09 & 0.00 & 0.00 & 0.04 & 0.00 & 0.87 & 0.00 & 0.00 & 0.12 \\ \hline
\multirow{4}{*}{\textbf{CIFAR10}} & \multirow{2}{*}{\textbf{ResNet20}} & \textbf{All} & 0.13 & 0.12 & 0.52 & 0.50 & 0.40 & 0.49 & 0.23 & 0.23 & 0.53 & 0.35 \\
 &  & \textbf{Type2 only} & 0.09 & 0.09 & 0.01 & 0.00 & 0.01 & 0.00 & 0.01 & 0.05 & 0.00 & 0.03 \\
 & \multirow{2}{*}{\textbf{VGG16}} & \textbf{All} & 0.17 & 0.17 & 0.45 & 0.46 & 0.36 & 0.46 & 0.26 & 0.22 & 0.44 & 0.33 \\
 &  & \textbf{Type2 only} & 0.09 & 0.10 & 0.00 & 0.00 & 0.01 & 0.00 & 0.07 & 0.07 & 0.00 & 0.04 \\ \hline
\multirow{4}{*}{\textbf{Traffic}} & \multirow{2}{*}{\textbf{LeNet5}} & \textbf{All} & 0.29 & 0.30 & 0.74 & 0.79 & 0.73 & 0.79 & 0.39 & 0.36 & 0.78 & 0.57 \\
 &  & \textbf{Type2 only} & 0.09 & 0.09 & 0.00 & 0.00 & 0.09 & 0.00 & 0.00 & 0.08 & 0.00 & 0.04 \\
 & \multirow{2}{*}{\textbf{VGG16}} & \textbf{All} & 0.17 & 0.17 & 0.55 & 0.56 & 0.49 & 0.56 & 0.16 & 0.18 & 0.54 & 0.37 \\
 &  & \textbf{Type2 only} & 0.10 & 0.09 & 0.00 & 0.00 & 0.13 & 0.00 & 0.00 & 0.08 & 0.00 & 0.04 \\ \hline
\multirow{4}{*}{\textbf{CIFAR100}} & \multirow{2}{*}{\textbf{ResNet50}} & \textbf{All} & 0.37 & 0.34 & 0.70 & 0.79 & 0.70 & 0.79 & 0.31 & 0.48 & 0.74 & 0.58 \\
 &  & \textbf{Type2 only} & 0.10 & 0.10 & 0.00 & 0.00 & 0.00 & 0.00 & 0.07 & 0.09 & 0.00 & 0.04 \\
 & \multirow{2}{*}{\textbf{DenseNet121}} & \textbf{All} & 0.20 & 0.18 & 0.60 & 0.63 & 0.61 & 0.65 & 0.16 & 0.39 & 0.58 & 0.45 \\
 &  & \textbf{Type2 only} & 0.08 & 0.08 & 0.00 & 0.00 & 0.00 & 0.00 & 0.06 & 0.10 & 0.00 & 0.04 \\ \hline
\multicolumn{2}{l}{\multirow{4}{*}{\textbf{Average}}} & \textbf{Type2 only} & 0.09 & 0.09 & 0.01 & 0.00 & 0.05 & 0.00 & 0.14 & 0.07 & 0.01 & 0.05 \\
\multicolumn{2}{l}{} & \textbf{Ori} & 0.14 & 0.61 & 0.64 & 0.69 & 0.93 & 0.70 & 0.76 & 0.55 & 0.67 & 0.63 \\
\multicolumn{2}{l}{} & \textbf{All} & 0.18 & 0.18 & 0.42 & 0.45 & 0.41 & 0.45 & 0.30 & 0.26 & 0.44 & 0.34 \\
\multicolumn{2}{l}{} & \textbf{Diff} & 22.28\% \contour{blue}{\textcolor{blue}{$\uparrow$}} & 71.40\% \contour{red}{\textcolor{red}{$\downarrow$}} & 35.85\% \contour{red}{\textcolor{red}{$\downarrow$}} & 55.58\% \contour{red}{\textcolor{red}{$\downarrow$}} & 35.16\% \contour{red}{\textcolor{red}{$\downarrow$}} & 35.78\% \contour{red}{\textcolor{red}{$\downarrow$}} & 59.86\% \contour{red}{\textcolor{red}{$\downarrow$}} & 52.33\% \contour{red}{\textcolor{red}{$\downarrow$}} & 34.55\% \contour{red}{\textcolor{red}{$\downarrow$}} & 39.80\% \contour{red}{\textcolor{red}{$\downarrow$}} \\ \hline
\end{tabular}
}
\end{table*}

Table~\ref{tab:type2_test_data} presents the results of fault detection, where $All$ is the TRC of the entire test set, and $Type2\ only$ shows the percentage of type2 faults that have been detected. We can see that MC, DeepGini, MaxP, and PRIMA cannot detect the \textit{Type2} faults where their $Type2\ only$ scores are nearly 0 across all the datasets and models. This means that these methods can not detect faults that are far away from the decision boundaries. Although other methods can detect some \textit{Type2} faults, most of them detect fewer than the random selection except TestRank. From the average results, we can see that compared to the TRC values on the original test data ($Ori$), the TRC values on \textit{Type2} data drop 39.80\% (46.70\% if without random selection). Similar to the analysis of \textit{Type1} data, we compare the performance of all methods on \textit{Type2} data and original test data using $t$-test. The results demonstrate that except for \textit{Random} and \textit{TestRank}, all methods perform significantly worse on \textit{Type2} data than on original test data (with a p-value \textless 0.05). Surprisingly, the intermediate output-based method, DSA has the greatest performance degradation. And the scores of $Type2\ only$ and $All$ are close to the scores achieved by random selection (0.09), which indicates that when facing the \textit{Type2} test data, DSA is completely ineffective. The potential reason could be that the distance map between the test data and the training data is significantly changed by the \textit{Type2} data and DSA is confused by the new distance map. The deeper analysis is an interesting research direction for our future work.

\noindent \\ 
{\framebox{\parbox{0.96\linewidth}{
\textbf{Answer to RQ3}: Existing fault detection methods cannot detect \textit{Type2} faults where the model has high confidence. On average, when facing \textit{Type2} test data, the test relative coverage scores achieved by test selection methods drop 39.80\%.}}}

\subsection{RQ4: Model Repair}

After finding the fault, the next step is to fix the fault. The most common approach the existing works~\cite{kim2019guiding, feng2020deepgini, ma2021test, 9402064, 9286133} apply is to use faults as \textit{patches} to repair~(by retraining) the pre-trained model. In this part, we study how \textit{Type1} and \textit{Type2} data affect the effectiveness of test selection-based model retraining. Table~\ref{tab:retrain} presents the detailed accuracy difference between pre-trained and re-trained models. First, it is interesting that when facing the \textit{Type1} test data, after model retraining, most of the models have accuracy degradation (55 out of 80 models). From the average results, we can see only ATS, DeepGini, MaxP, MCP, and MC can repair the pre-trained models but with negligible improvement (only by up to 0.90\% accuracy improvement). After significance analysis using $t$-test, we found that compared to retraining with original test data, all the methods (except for ATS) produce significantly worse models after retraining with \textit{Type1} data. On the other hand, for the \textit{Type2} test data, although most (70 out of 80) of the test selection methods can repair the pre-trained models and achieve positive accuracy improvement, compared to the performance of model repair on the original test data, the improvement is slight (i.e., only in 2 out of 80 cases, the results on \textit{Type2} data are better than the results on original data). However, different from the \textit{Type1} data, the $t$-test results demonstrate that only DSA produces significantly worse models after retraining on \textit{Type2} data. This phenomenon indicates that, for model retraining, \textit{Type1} data is more harmful than \textit{Type2} data. 

\begin{table*}[]
\centering
\caption{Accuracy difference (\%) and the variance (in brackets) between the pre-trained and retrained models (labeling budget 10\%). The best results among Ori, \textit{Type1}, and \textit{Type2} are highlighted using gray background. Models that have accuracy degradation after retraining are highlighted using orange background.}
\label{tab:retrain}
\resizebox{0.825\textwidth}{!}
{
\begin{tabular}{lllcccccccccc}
\hline
 &  &  & \textbf{Random} & \textbf{DSA} & \textbf{DeepGini} & \textbf{MCP} & \textbf{MC} & \textbf{MaxP} & \textbf{TestRank} & \textbf{PRIMA} & \textbf{ATS} & \textbf{Average} \\ \hline
 &  & \textbf{Ori} & \cellcolor[HTML]{EFEFEF}0.45 (0.05) & \cellcolor[HTML]{EFEFEF}0.59 (0.06) & \cellcolor[HTML]{EFEFEF}0.62 (0.07) & \cellcolor[HTML]{EFEFEF}0.67 (0.1) & \cellcolor[HTML]{EFEFEF}0.78 (0.06) & \cellcolor[HTML]{EFEFEF}0.74 (0.08) & \cellcolor[HTML]{EFEFEF}0.43 (0.06) & \cellcolor[HTML]{EFEFEF}0.55 (0.02) & \cellcolor[HTML]{EFEFEF}0.69 (0.04) & \cellcolor[HTML]{EFEFEF}0.61 (0.06) \\
 &  & \textbf{Type1} & \cellcolor[HTML]{F3E5AB}-1.60 (0.06) & \cellcolor[HTML]{F3E5AB}-1.66 (0.37) & 0.22 (0.08) & \cellcolor[HTML]{F3E5AB}-0.29 (0.51) & 0.25 (0.02) & 0.25 (0.08) & \cellcolor[HTML]{F3E5AB}-1.91 (0.28) & \cellcolor[HTML]{F3E5AB}-2.95 (0.82) & \cellcolor[HTML]{F3E5AB}-2.85 (0.89) & \cellcolor[HTML]{F3E5AB}-1.17 (0.34) \\
 & \multirow{-3}{*}{\textbf{LeNet1}} & \textbf{Type2} & 0.12 (0.2) & 0.24 (0.11) & 0.51 (0.12) & 0.52 (0.03) & 0.52 (0.09) & 0.53 (0.2) & 0.34 (0.11) & 0.48 (0.08) & 0.44 (0.16) & 0.41 (0.12) \\
 &  & \textbf{Ori} & \cellcolor[HTML]{EFEFEF}0.17 (0.06) & \cellcolor[HTML]{EFEFEF}0.94 (0.08) & \cellcolor[HTML]{EFEFEF}0.87 (0.22) & \cellcolor[HTML]{EFEFEF}0.91 (0.09) & \cellcolor[HTML]{EFEFEF}0.85 (0.08) & \cellcolor[HTML]{EFEFEF}0.81 (0.12) & \cellcolor[HTML]{EFEFEF}0.42 (0.02) & \cellcolor[HTML]{EFEFEF}0.48 (0.02) & \cellcolor[HTML]{EFEFEF}0.72 (0.05) & \cellcolor[HTML]{EFEFEF}0.68 (0.08) \\
 &  & \textbf{Type1} & \cellcolor[HTML]{F3E5AB}-1.00 (1.01) & \cellcolor[HTML]{F3E5AB}-0.33 (0.59) & \cellcolor[HTML]{F3E5AB}-0.03 (0.04) & \cellcolor[HTML]{F3E5AB}-0.64 (0.41) & 0.21 (0.08) & \cellcolor[HTML]{F3E5AB}-0.03 (0.07) & \cellcolor[HTML]{F3E5AB}-3.02 (0.55) & \cellcolor[HTML]{F3E5AB}-1.83 (0.7) & 0.33 (0.11) & \cellcolor[HTML]{F3E5AB}-0.71 (0.40) \\
\multirow{-6}{*}{\textbf{MNIST}} & \multirow{-3}{*}{\textbf{LeNet5}} & \textbf{Type2} & 0.04 (0.12) & \cellcolor[HTML]{F3E5AB}-0.11 (0.09) & 0.74 (0.13) & 0.70 (0.05) & 0.57 (0.20) & 0.80 (0.10) & 0.14 (0.16) & 0.29 (0.18) & 0.57 (0.30) & 0.42 (0.15) \\ \hline
 &  & \textbf{Ori} & \cellcolor[HTML]{EFEFEF}0.98 (0.30) & \cellcolor[HTML]{EFEFEF}2.19 (0.55) & \cellcolor[HTML]{EFEFEF}4.46 (0.34) & \cellcolor[HTML]{EFEFEF}3.81 (0.18) & \cellcolor[HTML]{EFEFEF}3.60 (0.27) & \cellcolor[HTML]{EFEFEF}4.27 (0.18) & \cellcolor[HTML]{EFEFEF}3.34 (0.32) & \cellcolor[HTML]{EFEFEF}1.15 (0.75) & \cellcolor[HTML]{EFEFEF}2.65 (0.20) & \cellcolor[HTML]{EFEFEF}2.94 (0.34) \\
 &  & \textbf{Type1} & \cellcolor[HTML]{F3E5AB}-2.29 (0.81) & \cellcolor[HTML]{F3E5AB}-3.18 (0.49) & \cellcolor[HTML]{F3E5AB}-2.97 (0.57) & \cellcolor[HTML]{F3E5AB}-0.79 (2.28) & 0.30 (0.89) & \cellcolor[HTML]{F3E5AB}-0.85 (0.49) & \cellcolor[HTML]{F3E5AB}-0.57 (0.46) & \cellcolor[HTML]{F3E5AB}-5.02 (0.15) & \cellcolor[HTML]{F3E5AB}-2.90 (1.38) & \cellcolor[HTML]{F3E5AB}-2.03 (0.83) \\
 & \multirow{-3}{*}{\textbf{LeNet5}} & \textbf{Type2} & \cellcolor[HTML]{F3E5AB}-0.50 (0.11) & \cellcolor[HTML]{F3E5AB}-0.01 (0.15) & 1.99 (0.32) & 1.74 (0.43) & 1.72 (0.28) & 1.46 (0.89) & 1.31 (0.32) & \cellcolor[HTML]{F3E5AB}-0.25 (0.95) & 0.67 (0.72) & 0.90 (0.46) \\
 &  & \textbf{Ori} & \cellcolor[HTML]{EFEFEF}0.15 (0.28) & \cellcolor[HTML]{EFEFEF}0.80 (0.20) & \cellcolor[HTML]{EFEFEF}1.33 (0.25) & \cellcolor[HTML]{EFEFEF}1.32 (0.31) & \cellcolor[HTML]{EFEFEF}0.95 (0.36) & \cellcolor[HTML]{EFEFEF}1.43 (0.12) & \cellcolor[HTML]{EFEFEF}0.61 (0.47) & \cellcolor[HTML]{EFEFEF}0.73 (0.30) & \cellcolor[HTML]{EFEFEF}0.96 (0.26) & \cellcolor[HTML]{EFEFEF}0.92 (0.28) \\
 &  & \textbf{Type1} & \cellcolor[HTML]{F3E5AB}-3.37 (1.38) & \cellcolor[HTML]{F3E5AB}-3.60 (0.74) & \cellcolor[HTML]{F3E5AB}-0.38 (0.63) & \cellcolor[HTML]{F3E5AB}-0.38 (0.66) & \cellcolor[HTML]{F3E5AB}-0.11 (0.21) & \cellcolor[HTML]{F3E5AB}-0.48 (0.08) & \cellcolor[HTML]{F3E5AB}-4.17 (0.31) & \cellcolor[HTML]{F3E5AB}-3.47 (0.74) & \cellcolor[HTML]{F3E5AB}-0.04 (0.80) & \cellcolor[HTML]{F3E5AB}-1.78 (0.62) \\
\multirow{-6}{*}{\textbf{SVHN}} & \multirow{-3}{*}{\textbf{ResNet20}} & \textbf{Type2} & \cellcolor[HTML]{F3E5AB}-0.02 (0.18) & \cellcolor[HTML]{F3E5AB}-0.37 (0.11) & 0.59 (0.21) & 0.43 (0.58) & 0.45 (0.52) & 0.71 (0.17) & \cellcolor[HTML]{F3E5AB}-0.16 (0.54) & 0.37 (0.15) & 0.56 (0.54) & 0.28 (0.33) \\ \hline
 &  & \textbf{Ori} & \cellcolor[HTML]{EFEFEF}0.78 (0.29) & \cellcolor[HTML]{EFEFEF}1.52 (0.35) & \cellcolor[HTML]{EFEFEF}3.64 (0.46) & \cellcolor[HTML]{EFEFEF}1.27 (0.26) & \cellcolor[HTML]{EFEFEF}2.84 (0.98) & \cellcolor[HTML]{EFEFEF}3.68 (0.62) & \cellcolor[HTML]{EFEFEF}0.93 (0.52) & \cellcolor[HTML]{EFEFEF}0.70 (0.36) & \cellcolor[HTML]{EFEFEF}3.15 (0.96) & \cellcolor[HTML]{EFEFEF}2.06 (0.53) \\
 &  & \textbf{Type1} & \cellcolor[HTML]{F3E5AB}-1.66 (2.04) & \cellcolor[HTML]{F3E5AB}-3.76 (1.50) & \cellcolor[HTML]{F3E5AB}-0.71 (0.58) & \cellcolor[HTML]{F3E5AB}-0.63 (0.61) & \cellcolor[HTML]{F3E5AB}-3.50 (2.44) & \cellcolor[HTML]{F3E5AB}-1.42 (0.84) & \cellcolor[HTML]{F3E5AB}-2.77 (2.46) & \cellcolor[HTML]{F3E5AB}-4.54 (0.41) & \cellcolor[HTML]{F3E5AB}-0.59 (0.10) & \cellcolor[HTML]{F3E5AB}-2.17 (1.22) \\
 & \multirow{-3}{*}{\textbf{ResNet20}} & \textbf{Type2} & \cellcolor[HTML]{F3E5AB}-0.69 (0.56) & \cellcolor[HTML]{F3E5AB}-1.25 (1.30) & 0.34 (0.35) & \cellcolor[HTML]{F3E5AB}-0.06 (0.52) & 0.98 (0.12) & 0.60 (0.60) & \cellcolor[HTML]{F3E5AB}-0.12 (0.70) & \cellcolor[HTML]{F3E5AB}-1.09 (0.79) & 0.62 (0.43) & \cellcolor[HTML]{F3E5AB}-0.08 (0.60) \\
 &  & \textbf{Ori} & \cellcolor[HTML]{EFEFEF}2.36 (0.17) & \cellcolor[HTML]{EFEFEF}3.83 (0.08) & \cellcolor[HTML]{EFEFEF}6.80 (0.19) & \cellcolor[HTML]{EFEFEF}5.18 (0.13) & \cellcolor[HTML]{EFEFEF}6.56 (0.17) & \cellcolor[HTML]{EFEFEF}6.81 (0.18) & \cellcolor[HTML]{EFEFEF}1.98 (0.11) & \cellcolor[HTML]{EFEFEF}2.17 (0.3) & \cellcolor[HTML]{EFEFEF}4.67 (0.11) & \cellcolor[HTML]{EFEFEF}4.48 (0.13) \\
 &  & \textbf{Type1} & 0.75 (0.11) & 0.78 (0.12) & 2.09 (0.46) & 1.83 (0.04) & 1.40 (0.07) & 1.12 (0.15) & \cellcolor[HTML]{F3E5AB}-1.57 (0.23) & \cellcolor[HTML]{F3E5AB}-0.50 (0.06) & 3.68 (0.06) & 1.06 (0.15) \\
\multirow{-6}{*}{\textbf{CIFAR10}} & \multirow{-3}{*}{\textbf{VGG16}} & \textbf{Type2} & 1.34 (0.04) & 1.00 (0.2) & 4.07 (0.04) & 3.10 (0.1) & 4.04 (0.03) & 4.06 (0.02) & 1.91 (0.10) & 1.54 (0.14) & 4.08 (0.07) & 2.79 (0.05) \\ \hline
 &  & \textbf{Ori} & \cellcolor[HTML]{EFEFEF}6.41 (0.10) & \cellcolor[HTML]{EFEFEF}7.25 (0.19) & \cellcolor[HTML]{EFEFEF}10.24 (0.74) & \cellcolor[HTML]{EFEFEF}6.79 (0.55) & \cellcolor[HTML]{EFEFEF}10.16 (0.42) & \cellcolor[HTML]{EFEFEF}9.85 (0.24) & \cellcolor[HTML]{EFEFEF}2.15 (0.31) & \cellcolor[HTML]{EFEFEF}6.97 (0.66) & \cellcolor[HTML]{EFEFEF}7.68 (0.87) & \cellcolor[HTML]{EFEFEF}7.50 (0.45) \\
 &  & \textbf{Type1} & 1.96 (0.60) & 1.58 (1.82) & 4.63 (0.71) & 4.30 (0.62) & 4.55 (0.86) & 4.01 (1.02) & \cellcolor[HTML]{F3E5AB}-0.58 (0.70) & 2.02 (0.06) & 5.68 (0.36) & 3.13 (0.75) \\
 & \multirow{-3}{*}{\textbf{LeNet5}} & \textbf{Type2} & 1.90 (0.00) & 3.09 (1.01) & 7.27 (0.10) & 6.49 (0.74) & 6.85 (0.19) & 6.58 (0.07) & 3.26 (0.10) & 3.21 (0.23) & 6.56 (0.59) & 5.02 (0.34) \\
 &  & \textbf{Ori} & \cellcolor[HTML]{EFEFEF}4.52 (0.28) & \cellcolor[HTML]{EFEFEF}5.36 (0.65) & \cellcolor[HTML]{EFEFEF}6.10 (0.18) & \cellcolor[HTML]{EFEFEF}4.43 (1.10) & \cellcolor[HTML]{EFEFEF}5.30 (0.68) & \cellcolor[HTML]{EFEFEF}5.09 (0.49) & \cellcolor[HTML]{EFEFEF}2.19 (0.03) & 2.86 (0.56) & 3.97 (0.72) & \cellcolor[HTML]{EFEFEF}4.42 (0.52) \\
 &  & \textbf{Type1} & 1.02 (0.64) & 0.59 (2.14) & 3.41 (0.28) & 3.40 (0.20) & 3.01 (0.27) & 2.61 (0.92) & 0.27 (0.64) & \cellcolor[HTML]{F3E5AB}-0.64 (0.83) & \cellcolor[HTML]{EFEFEF}4.35 (0.96) & 2.00 (0.76) \\
\multirow{-6}{*}{\textbf{Traffic}} & \multirow{-3}{*}{\textbf{VGG16}} & \textbf{Type2} & 3.83 (0.29) & 0.67 (0.21) & 3.99 (0.43) & 3.89 (0.41) & 4.24 (0.66) & 3.64 (0.78) & 1.26 (0.83) & \cellcolor[HTML]{EFEFEF}2.88 (0.61) & 4.31 (0.26) & 3.19 (0.50) \\ \hline
 &  & \textbf{Ori} & \cellcolor[HTML]{EFEFEF}1.35 (1.30) & 0.27 (0.86) & \cellcolor[HTML]{EFEFEF}1.64 (0.60) & \cellcolor[HTML]{EFEFEF}3.18 (3.26) & \cellcolor[HTML]{EFEFEF}3.27 (2.04) & 1.06 (0.77) & \cellcolor[HTML]{EFEFEF}2.37 (0.61) & \cellcolor[HTML]{EFEFEF}1.58 (0.71) & 1.15 (1.60) & \cellcolor[HTML]{EFEFEF}1.76 (1.31) \\
 &  & \textbf{Type1} & \cellcolor[HTML]{F3E5AB}-1.18 (1.65) & \cellcolor[HTML]{F3E5AB}-1.45 (1.75) & \cellcolor[HTML]{F3E5AB}-1.51 (2.55) & 0.11 (0.05) & \cellcolor[HTML]{F3E5AB}-0.02 (1.18) & 0.59 (1.39) & \cellcolor[HTML]{F3E5AB}-1.66 (0.97) & \cellcolor[HTML]{F3E5AB}-0.95 (0.97) & 1.09 (0.44) & \cellcolor[HTML]{F3E5AB}-0.55 (1.22) \\
 & \multirow{-3}{*}{\textbf{ResNet50}} & \textbf{Type2} & 0.57 (2.81) & \cellcolor[HTML]{EFEFEF}1.19 (2.13) & 1.36 (1.71) & 1.72 (0.58) & 0.98 (1.61) & \cellcolor[HTML]{EFEFEF}2.12 (0.59) & 0.87 (1.53) & 1.25 (0.45) & \cellcolor[HTML]{EFEFEF}1.63 (1.58) & 1.30 (1.44) \\
 &  & \textbf{Ori} & \cellcolor[HTML]{EFEFEF}0.68 (0.14) & \cellcolor[HTML]{EFEFEF}2.75 (0.26) & \cellcolor[HTML]{EFEFEF}8.76 (0.32) & 0.88 (0.17) & \cellcolor[HTML]{EFEFEF}9.01 (0.28) & \cellcolor[HTML]{EFEFEF}9.23 (0.33) & \cellcolor[HTML]{EFEFEF}0.43 (0.07) & \cellcolor[HTML]{EFEFEF}1.75 (0.24) & \cellcolor[HTML]{EFEFEF}4.53 (0.10) & \cellcolor[HTML]{EFEFEF}4.22 (0.21) \\
 &  & \textbf{Type1} & \cellcolor[HTML]{F3E5AB}-3.26 (0.12) & \cellcolor[HTML]{F3E5AB}-3.27 (0.17) & \cellcolor[HTML]{F3E5AB}-1.60 (0.03) & 2.97 (0.11) & 0.49 (0.08) & \cellcolor[HTML]{F3E5AB}-1.37 (0.16) & \cellcolor[HTML]{F3E5AB}-4.48 (0.16) & \cellcolor[HTML]{F3E5AB}-3.11 (0.19) & 1.17 (0.31) & \cellcolor[HTML]{F3E5AB}-1.39 (0.15) \\
\multirow{-6}{*}{\textbf{CIFAR100}} & \multirow{-3}{*}{\textbf{DenseNet101}} & \textbf{Type2} & \cellcolor[HTML]{F3E5AB}-0.69 (0.37) & \cellcolor[HTML]{F3E5AB}-0.93 (0.19) & \cellcolor[HTML]{F3E5AB}4.33 (0.28) & \cellcolor[HTML]{EFEFEF}4.16 (0.24) & 4.43 (0.21) & 4.58 (0.25) & \cellcolor[HTML]{F3E5AB}-0.75 (0.06) & 0.71 (0.27) & 3.88 (0.35) & 2.19 (0.25) \\ \hline
\multicolumn{2}{l}{} & \textbf{Ori} & \cellcolor[HTML]{EFEFEF}1.62 (0.30) & \cellcolor[HTML]{EFEFEF}2.32 (0.33) & \cellcolor[HTML]{EFEFEF}4.04 (0.34) & \cellcolor[HTML]{EFEFEF}2.58 (0.62) & \cellcolor[HTML]{EFEFEF}3.94 (0.53) & \cellcolor[HTML]{EFEFEF}3.91 (0.31) & \cellcolor[HTML]{EFEFEF}1.35 (0.25) & \cellcolor[HTML]{EFEFEF}1.72 (0.36) & \cellcolor[HTML]{EFEFEF}2.74 (0.49) & \cellcolor[HTML]{EFEFEF}2.69 (0.25) \\
\multicolumn{2}{l}{} & \textbf{Type1} & \cellcolor[HTML]{F3E5AB}-0.97 (0.55) & \cellcolor[HTML]{F3E5AB}-1.30 (0.97) & 0.29 (0.59) & 0.90 (0.52) & 0.60 (0.61) & 0.40 (0.49) & \cellcolor[HTML]{F3E5AB}-1.86 (0.68) & \cellcolor[HTML]{F3E5AB}-1.91 (0.89) & 0.90 (0.54) & \cellcolor[HTML]{F3E5AB}-0.33 (0.64) \\
\multicolumn{2}{l}{\multirow{-3}{*}{\textbf{Average}}} & \textbf{Type2} & 0.54 (0.47) & 0.32 (0.53) & 2.29 (0.37) & 2.06 (0.36) & 2.25 (0.39) & 2.28 (0.37) & 0.73 (0.45) & 0.85 (0.38) & 2.12 (0.50) & 1.49 (0.42) \\ \hline
\end{tabular}
}
\end{table*}

\begin{wrapfigure}[12]{c}{0.55\linewidth}
 \centering
 \includegraphics[width =\linewidth]{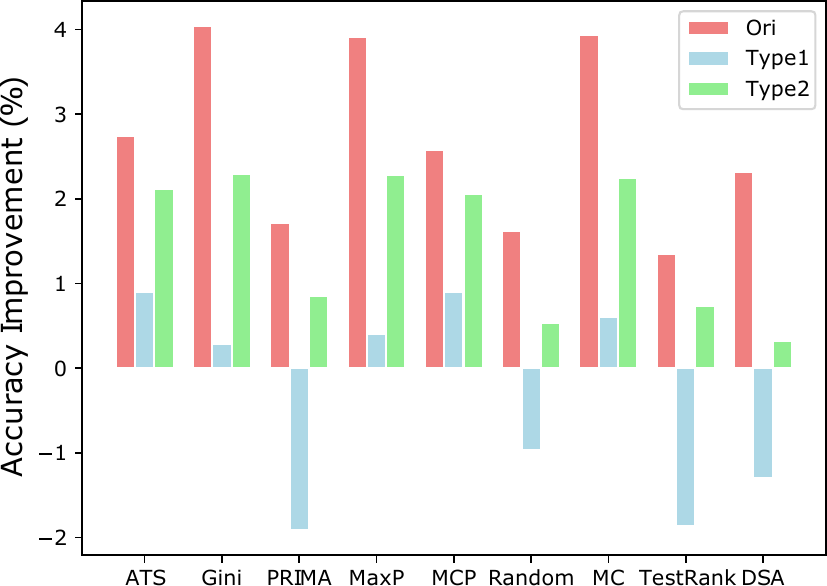}
 \caption{Repair results by different test selection methods.}\label{fig:rq3}
\end{wrapfigure}

Figure~\ref{fig:rq3} depicts the average accuracy improvements of model repair on all datasets and models. The results clearly show that when the test set contains \textit{Type1} and \textit{Type2} data, the effectiveness of test selection-based model repair is worse than when there are only original test data. Considering different test selection methods, we can see that the accuracy improvement by PRIMA and TestRank is always lower than by random selection. Note that in their original work, they did not check the effectiveness of test selection-based model repair. This reminds us that when proposing new fault detection target test selection methods, we should also explore whether the revealed faults are useful for repairing the model or not. For DSA, although it achieves better results than random selection on the original test data, its performance on the \textit{Type1} and \textit{Type2} test data are worse than random selection. This indicates, for model repair, that DSA is only suitable for standard test data that share the same characteristics as the training set.

\noindent \\ 
{\framebox{\parbox{0.96\linewidth}{
\textbf{Answer to RQ4}: \textit{Type1} and \textit{Type2} test data harm the performance of selection-based model repair. Given these two types of data, DSA, PRIMA, and TestRank achieve worse repair results than random selection. Especially, more than half (55 out of 80) of repaired models occur accuracy degradation with \textit{Type1} data.}}}

\subsection{RQ5: Performance Estimation}

Finally, we explore how the choice of intermediate output affects the effectiveness of performance estimation methods. Table~\ref{tab:acc_estimation} shows the frequency of test selection methods achieving the best results over different labeling budgets. The first two columns of values are the comparison between different intermediate layers in the same test selection method. The last three columns are the comparison between different methods. We can see that, first, by the same method, the average results suggest the second-last hidden layer as the best choice for both CES and PACE methods. However, when we check the results of each dataset and model, it is difficult to decide which layer should be used. For example, for CES, layer-2 is the best choice for MNIST-LeNet5, but layer-1 is the best one for SVHN-LeNet5. Similar to CES, for PACE, layer-3 achieves the best results on MNIST-LeNet5 while layer-2 is the best for SVHN-LeNet5. This means the choice of an intermediate layer highly impacts the results of these two methods and there is no clear conclusion on which layer we should choose when using different models. Then, if we compare different methods, the average results demonstrate that only CES with the outputs from layer-2 (0.44) can significantly outperform the random selection (0.38), which means if we choose an unsuitable intermediate layer, the results achieved by the well-designed methods are worse than random selection.

\begin{table}[]
\centering
\caption{Frequency of each test selection method achieving the top-1 performance using different intermediate outputs. \textit{layer-N} means the output is from the last Nth layer.}
\label{tab:acc_estimation}
\resizebox{.85\columnwidth}{!}
{
\begin{tabular}{lllcc|ccc}
\hline
\multicolumn{3}{l}{} & \multicolumn{1}{l}{\textbf{CES}} & \multicolumn{1}{l|}{\textbf{PACE}} & \multicolumn{1}{l}{\textbf{CES}} & \multicolumn{1}{l}{\cellcolor[HTML]{FFFFFF}\textbf{PACE}} & \multicolumn{1}{l}{\cellcolor[HTML]{FFFFFF}\textbf{Random}} \\ \hline
 &  & \textbf{layer-1} & 0.29 & 0.00 & 0.64 & 0.00 & 0.36 \\
 &  & \textbf{layer-2} & 0.36 & 0.21 & 0.29 & 0.64 & 0.07 \\
 & \multirow{-3}{*}{\textbf{LeNet1}} & \textbf{layer-3} & 0.36 & 0.79 & 0.00 & 0.93 & 0.07 \\
 &  & \textbf{layer-1} & 0.07 & 0.29 & 0.21 & 0.29 & 0.50 \\
 &  & \textbf{layer-2} & 0.64 & 0.29 & 0.29 & 0.29 & 0.43 \\
\multirow{-6}{*}{\textbf{MNSIT}} & \multirow{-3}{*}{\textbf{LeNet5}} & \textbf{layer-3} & 0.29 & 0.43 & 0.14 & 0.29 & 0.57 \\ \hline
 &  & \textbf{layer-1} & 0.57 & 0.00 & 0.71 & 0.00 & 0.29 \\
 &  & \textbf{layer-2} & 0.14 & 0.79 & 0.57 & 0.21 & 0.21 \\
 & \multirow{-3}{*}{\textbf{LeNet5}} & \textbf{layer-3} & 0.29 & 0.21 & 0.43 & 0.36 & 0.21 \\
 &  & \textbf{layer-1} & 0.57 & 0.00 & 0.71 & 0.00 & 0.29 \\
 &  & \textbf{layer-2} & 0.14 & 0.57 & 0.36 & 0.29 & 0.36 \\
\multirow{-6}{*}{\textbf{SVHN}} & \multirow{-3}{*}{\textbf{ResNet20}} & \textbf{layer-3} & 0.29 & 0.29 & 0.43 & 0.29 & 0.29 \\ \hline
 &  & \textbf{layer-1} & 0.21 & 0.00 & 0.64 & 0.00 & 0.36 \\
 &  & \textbf{layer-2} & 0.21 & 0.14 & 0.64 & 0.07 & 0.29 \\
 & \multirow{-3}{*}{\textbf{ResNet20}} & \textbf{layer-3} & 0.57 & 0.86 & 0.43 & 0.57 & 0.00 \\
 &  & \textbf{layer-1} & 0.29 & 0.00 & 0.57 & 0.00 & 0.43 \\
 &  & \textbf{layer-2} & 0.50 & 0.93 & 0.71 & 0.00 & 0.29 \\
\multirow{-6}{*}{\textbf{CIFAR10}} & \multirow{-3}{*}{\textbf{VGG16}} & \textbf{layer-3} & 0.21 & 0.07 & 0.50 & 0.00 & 0.50 \\ \hline
 &  & \textbf{layer-1} & 0.21 & 1.00 & 0.57 & 0.00 & 0.43 \\
 &  & \textbf{layer-2} & 0.50 & 0.00 & 0.50 & 0.00 & 0.50 \\
 & \multirow{-3}{*}{\textbf{LeNet5}} & \textbf{layer-3} & 0.29 & 0.00 & 0.64 & 0.00 & 0.36 \\
 &  & \textbf{layer-1} & 0.57 & 0.36 & 0.43 & 0.00 & 0.57 \\
 &  & \textbf{layer-2} & 0.21 & 0.36 & 0.36 & 0.00 & 0.64 \\
\multirow{-6}{*}{\textbf{Traffic}} & \multirow{-3}{*}{\textbf{VGG16}} & \textbf{layer-3} & 0.21 & 0.29 & 0.14 & 0.00 & 0.86 \\ \hline
 &  & \textbf{layer-1} & 0.36 & 0.79 & 0.14 & 0.00 & 0.86 \\
 &  & \textbf{layer-2} & 0.43 & 0.07 & 0.29 & 0.00 & 0.71 \\
 & \multirow{-3}{*}{\textbf{ResNet50}} & \textbf{layer-3} & 0.21 & 0.14 & 0.21 & 0.00 & 0.79 \\
 &  & \textbf{layer-1} & 0.14 & 0.00 & 0.21 & 0.00 & 0.79 \\
 &  & \textbf{layer-2} & 0.50 & 0.93 & 0.36 & 0.36 & 0.29 \\
\multirow{-6}{*}{\textbf{CIFAR100}} & \multirow{-3}{*}{\textbf{DenseNet121}} & \textbf{layer-3} & 0.36 & 0.07 & 0.50 & 0.21 & 0.29 \\ \hline
\multicolumn{2}{l}{} & \textbf{layer-1} & 0.33 & 0.24 & 0.49 & 0.03 & 0.49 \\
\multicolumn{2}{l}{} & \textbf{layer-2} & 0.36 & 0.43 & 0.44 & 0.19 & 0.38 \\
\multicolumn{2}{l}{\multirow{-3}{*}{\textbf{Average}}} & \textbf{layer-3} & 0.31 & 0.31 & 0.34 & 0.26 & 0.39 \\ \hline
\end{tabular}
}
\end{table}

\begin{figure}[!h]
	\centering
	\includegraphics[width=0.4\textwidth]{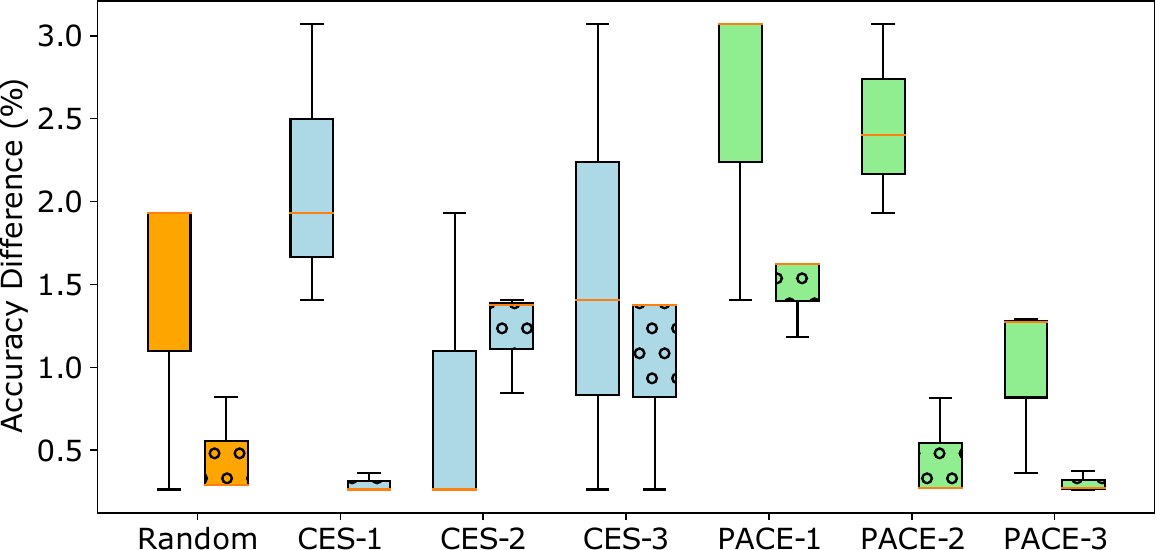}
	\caption{Results of accuracy estimation with labeling budgets 50 and 180 (filled with circle).}
	\label{fig:rq4}
    \vspace{-5mm}
\end{figure}

In addition, we check the effectiveness of performance estimation under different labeling budgets. Figure~\ref{fig:rq4} depicts the results of these three methods on labeling budgets 50 (minimum) and 180 (maximum). The results clearly show that when the labeling budget is 50, the second last hidden is more suitable for CES, but when the labeling budget is 180, the last hidden layer is the best. Besides, when selecting an unsuitable intermediate layer, the estimated results are worse than the random selection, e.g., CES-2 and PACE-1.

\noindent \\
{\framebox{\parbox{0.96\linewidth}{
\textbf{Answer to RQ5}: Performance estimation methods are sensitive to the choice of intermediate outputs. Those methods perform worse than random selection when an unsuitable layer is chosen. Unfortunately, it is difficult to determine the best layer since there is no clear conclusion across different datasets and models.}}}

\section{Discussion}
\label{sec:discussion}

\subsection{Guidance}

Here, we provide guidelines for proposing and evaluating test selection methods:

\begin{enumerate}[leftmargin=*]

\item From RQ2 and RQ3, we can see final output-based fault detection methods have critical constraints. To propose a fault detection method, only using the output probability is insufficient. The method can be easily fooled by uncommon data (like \textit{Type1} and \textit{Type2} data). It is better to avoid only relying on the output probability, e.g., combining the output and features from the input itself.   

\item For learning-based methods (e.g., TestRank), the learning models (e.g., simple GNN used in TestRank) should be evaluated first to check if they can distinguish features of any type of faults (not only the faults in the original test data but also harder data like \textit{Type1} and \textit{Type2}) and correctly classified data.

\item RQ4 shows that when there are \textit{Type1} data in the candidate data, retraining is ineffective and the computing resource is wasted. Thus, before repairing models via retraining, it is suggested to use out-of-distribution detection techniques~\cite{msp2017,ma18kim} to check the distribution of test data. 

\item RQ5 demonstrates that it is difficult to choose the appropriate layer for the use of existing performance estimation methods. Therefore, when proposing performance estimation methods, a layer-selection solution should be developed along with the method. It is impractical to use all intermediate outputs due to the high complexity.

\end{enumerate}

\subsection{Threats to Validity}

The \textbf{external threat} lies in the considered test selection methods, datasets, and models. For test selection methods, we collect methods that are specifically designed for test selection. Others such as neural coverage methods and active learning methods are not considered since they are proposed for different targets. For datasets and models, we use 5 datasets spanning from digit recognition to more practical traffic sign classification. And for each dataset, we include two model architectures, which can, to some extent, alleviate the model dependency. The \textbf{internal threat} can be the implementation of test selection methods and the GA-based test generation algorithm. All implementations of selection methods are modified from the official projects provided by corresponding authors. The implementation of GA-based test generation is also based on an existing work~\cite{zhang2020towards}. The \textbf{construct threat} can be the configuration of test selection methods and test generation process. For test selection methods, we follow their original papers and use default settings. For the configuration of the test generation process, we set parameters adaptively given the dataset and model. Note that our target is not to find a perfect parameter setting to attack test selection methods, instead, we focus on proving that non-robust features are easy to reveal.

\section{Conclusion}
\label{sec:conclusion}
We identified and systematically assessed three types of pitfalls in existing test selection methods for reliable testing of deep learning (DL)-based systems. Via an exploratory study, we found that methods for fault detection skip faults (up to 91\%) if a DL model is confident in the misclassification and introduce fake faults (up to 100\%) if a model has low confidence. In addition, selected data can degrade the accuracy when repairing models (e.g., 5.02\% degradation). On the other hand, methods for performance estimation fail to defeat the simplest random selection when using an inappropriate intermediate layer. Ultimately, we provide actionable guidelines on how to mitigate pitfalls when applying existing selection methods and avoid pitfalls when developing new ones.

\bibliographystyle{IEEEtran}
\bibliography{IEEEabrv,sample-base}

\end{document}